\documentclass[journal]{IEEEtran}

\usepackage{graphicx}
\usepackage{subfigure}
\usepackage{url}
\usepackage{times}
\usepackage{amssymb}
\usepackage{array}
\usepackage[ruled,lined]{algorithm2e}
\usepackage{amsmath}
\usepackage{booktabs}
\usepackage{multirow}
\usepackage{cite}
\usepackage{enumerate}

\begin{document}

\title{ Attention: A Big Surprise for Cross-Domain Person Re-Identification}   

\author{Haijun Liu, Jian Cheng, Shiguang Wang and Wen Wang

\thanks{H. Liu, J. Cheng, S. Wang and W. Wang are with the School of Information and Communication Engineering, University of Electronic Science and Technology of China, Chengdu, Sichuan, China, 611731.(Corresponding author: chengjian@uestc.edu.cn)}

}

\markboth{}
{Liu \MakeLowercase{\textit{et al.}}: ADFL for person Re-ID}

\maketitle

\begin{abstract}
In this paper, we focus on model generalization and adaptation for cross-domain person re-identification (Re-ID).
Unlike existing cross-domain Re-ID methods, leveraging the auxiliary information of those unlabeled target-domain data, we aim at enhancing the model generalization and adaptation by  discriminative feature learning, and directly exploiting a pre-trained model to new domains (datasets) without any utilization of the information from target domains.
To address the discriminative feature learning problem, we surprisingly find that simply introducing the attention mechanism to adaptively extract the person features for every domain is of great effectiveness.
We adopt two popular type of attention mechanisms, long-range dependency based attention and direct generation based attention. Both of them can perform the attention via spatial or channel dimensions alone, even the combination of spatial and channel dimensions. The outline of different attentions are well illustrated.
Moreover, we also incorporate the attention results into the final output of model through skip-connection to improve the features with both high and middle level semantic visual information.
In the manner of directly exploiting a pre-trained model to new domains, the attention incorporation method truly could enhance the model generalization and adaptation to perform the cross-domain person Re-ID. We conduct extensive experiments between three large datasets, Market-1501, DukeMTMC-reID and MSMT17. Surprisingly, introducing only attention can achieve state-of-the-art performance, even much better than those cross-domain Re-ID methods utilizing auxiliary information from the target domain.
\end{abstract}


\section{Introduction}
\label{sec:intro}

Person re-identification (Re-ID), aims at retrieving a person of interest across multiple non-overlapping cameras deployed at different locations via images matching \cite{ding2018feature}. It is a long-lasting research topic due to the wide range of applications in intelligent video surveillance \cite{wei2018glad,wang2015zero}. Re-ID is a challenging task because of significant pose-variations, varying illumination conditions, frequent human occlusions, background clutter and different camera views. All of those conditions give rise to the notorious image matching misalignment challenge in cross-view Re-ID, leading to the difficulty of extracting discriminative person features. However, extracting those subtle features that fully characterize the person, at the same time distinguish from other people, is not straightforward due to the existence of small inter-class variations and large intra-class differences. Therefore, how to obtain discriminative person features is crucial for person Re-ID.

\begin{figure}
  \centering
  \includegraphics[width=85mm]{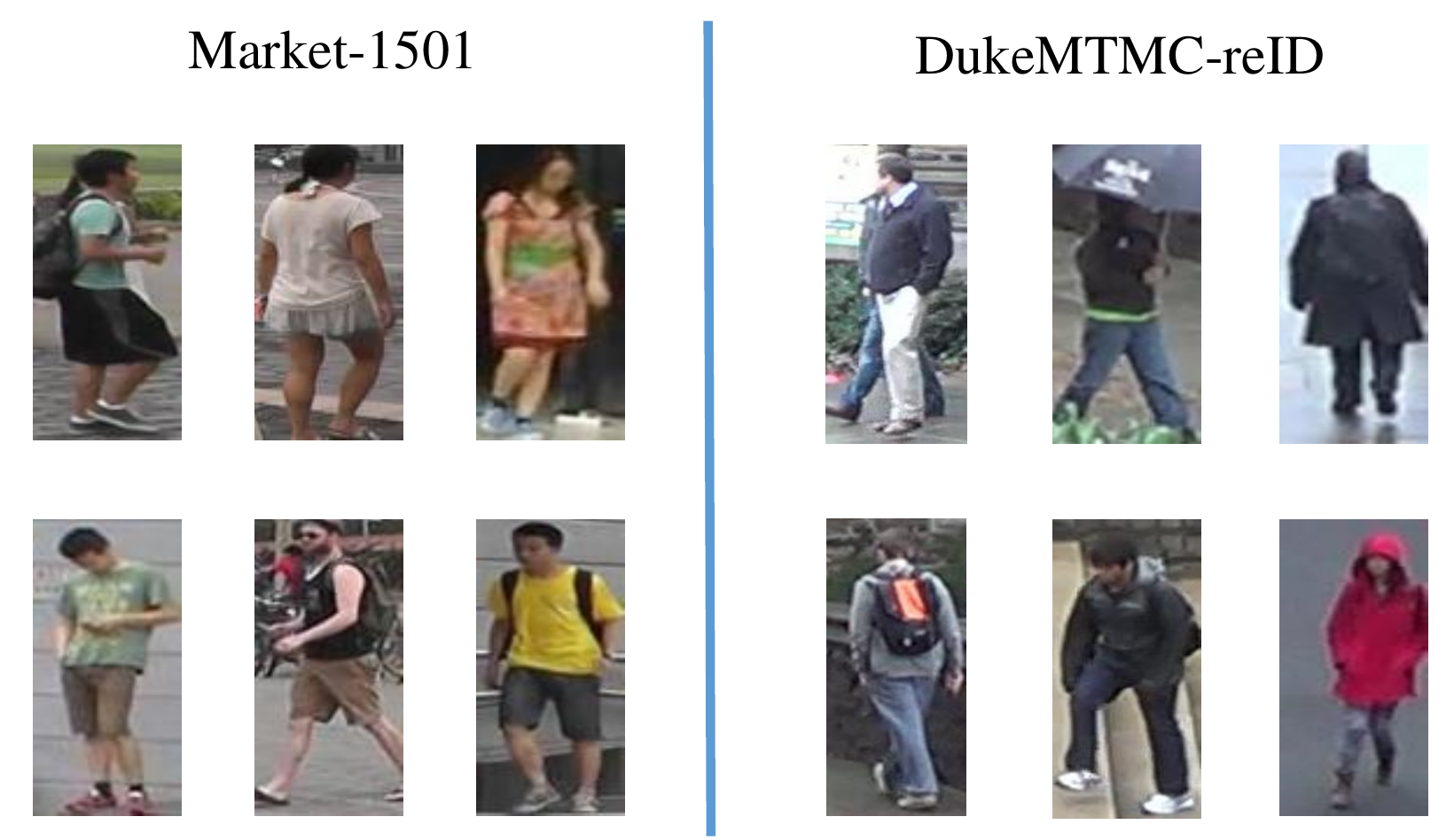}
  \caption{Example images for illustration of the different data distributions of the two Re-ID datasets, Market-1501 and DukeMTMC-reID. The identities in Market-1501 \cite{zheng2015scalable} always wear shorts, while those in DukeMTMC-reID \cite{ristani2016performance} are usually with coats and trousers.}
  \label{fig:images}
 \end{figure}

Person Re-ID can be considered as a fine-grained classification task, also is a zero-shot setting task, where the test identities are never seen during training. Even in some practical scenarios, Re-ID is still a cross-domain task, where training and testing are under different domains (datasets) with different camera networks; the data distributions between training and testing domains are also apparent different, e.g., Market-1501 \cite{zheng2015scalable} contains identities always wearing shorts, while DukeMTMC-reID \cite{ristani2016performance} consists of pedestrians usually with coats and trousers, as shown in Fig. \ref{fig:images}.
All of these discrepancies put forward higher demands for Re-ID model to adaptively extract the discriminative and robust person features.

With the prosperity of deep learning, deep convolutional neural network (CNN) has dominated this community for extracting representations of person images with better discrimination and robustness, and significantly leverages the performance of person Re-ID tasks.  The main superiority of CNN is that CNN can optimize the network parameters by arranging the visual feature extraction, metric learning and classification in an end-to-end learning manner.
From the perspective of feature extraction, those deep learning based methods for person Re-ID can be divided into two main categories.

The first category is the global feature learning, an intuitive method to extract person features by deep convolutional neural networks from the whole body on images \cite{zheng2016person,xiao2017joint,sun2017svdnet,bai2017scalable,wang2018resource,chang2018multi,xiong2018towards}.
The global feature learning aims to capture the most salient clues of appearance to describe identities and distinguish them from other people. Most of the state-of-the-art CNN based person Re-ID methods are adopting the pre-trained CNN models (e.g., ResNet \cite{he2016deep} model) on ImageNet and fine-tuning them on the person Re-ID datasets under the supervision of different losses (e.g., softmax and triplet loss \cite{hermans2017defense}).
The second category is the partial feature learning. The latest state-of-the-art on Re-ID benchmarks \cite{kalayeh2018human,sarfraz2018pose,suh2018part,xu2018attention,sun2018beyond,wang2018learning,fu2018horizontal,liu20183,li2018harmonious,zheng2018re,wang2018mancs} are almost all  achieved with deep-learned part features, which confirms that locating significant body parts from the whole images to represent the local features of person is an effective approach for boosting the performance of Re-ID. The deep-learned part features also can be adopted as an important complement for global features.

In the single-domain setting, training and testing on single dataset, person Re-ID has achieved significant performance. However, in the cross-domain setting, training on one dataset while testing on another dataset, those Re-ID methods directly exploiting a pre-trained model to new domains (datasets) were always along with huge performance drop. More and more researchers are leveraging the auxiliary information of those unlabeled target-domain data to improve the cross-domain person Re-ID \cite{wang2018transferable,deng2018image,wei2018person,lv2018unsupervised,huang2018eanet}.
However, using auxiliary information of those unlabeled target-domain data always faces more tasks and high complexities, such as, pose estimation \cite{huang2018eanet} and image translation \cite{deng2018image,wei2018person}. \textbf{Therefore, we rise to the challenge focusing on performing the cross-domain person Re-ID by directly exploiting a pre-trained model to new domains, without any information of the target domain.}

In this case, the most challenge point is how to improve the generalization and adaptation of the pre-trained model. Namely, how to make sure that the model pre-trained on one dataset still can extract the discriminative features on another dataset. 
We argue that those models essentially with the ability of adaptively extracting the discriminative features on different datasets are with high generalization and adaptation.
Therefore, to address this problem, we propose to introduce the attention mechanism for discriminative feature learning to enhance the model generalization and adaptation in cross-domain person Re-ID.

It is well known that the attention mechanism plays an important role in human visual perception system \cite{itti1998model,rensink2000dynamic,corbetta2002control}. Human visual perception does not process the whole image at once, but exploits a sequence of partial glimpses and selectively focuses on salient parts to well capture the visual information \cite{larochelle2010learning}. It is one kind of partial feature learning approach. Attention mechanism dynamically focuses on salient parts according to the detailed content of each image, which can make a high-level information integration to emphasise the salient aspects. This characteristic can contribute for enhancing the model generalization and adaptation for cross-domain person Re-ID.

In this paper, we address the model generalization and adaptation under single-dataset training setting for cross-domain person Re-ID, by focusing on adaptively extracting the discriminative person features. Attention mechanism is introduced for discriminative feature learning to perform the cross-domain Re-ID task.
We adopt two popular type of attention mechanisms, long-range dependency based attention and direct generation based attention. Both of them can perform the attention via spatial or channel dimensions alone, even the combination of spatial and channel dimensions. We also illustrate the structures of different attentions.
Based on a strong baseline, the attention modules are incorporated to improve the performance, especially the cross-domain person Re-ID by directly exploiting a pre-trained CNN model to new domains.
In summary, our paper has the following main contributions.
\begin{enumerate}
\item We implement a strong baseline, based on ResNet50 model, with some empirically architecture modifications and training strategies.
\item We introduce two type of attention modules, type I: long-range dependency based attention and type II: direct generation based attention. Both of them leverage the attention mechanism from perspective of spatial and channel dimensions. The possible arrangements to combine the spatial and channel attention modules are explored.
\item By a simple way to incorporate attention, we surprisingly find the great effectiveness of attention for enhancing the model generalization and adaptation.
\item By directly exploiting a pre-trained CNN model to new domains, the effectiveness of attention for enhancing the model generalization and adaptation, is demonstrated through the excellent performances on three Re-ID datasets, especially those cross-domain Re-ID experiments.
\end{enumerate}

\section{Related works}
\label{sec:relate}
In this section, we will briefly review some related works for person Re-ID from the following aspects, discriminative feature learning, attention and cross-domain.

\textbf{Discriminative feature learning:}
Recently the performance of deep person Re-ID has been pushed to a new level. Among all of these strategies, focusing on the local features from parts of person images may be the most effective one to handle the misalignment of person images.
To accurately locate the body parts with semantics, pose estimation methods \cite{cao2017realtime,xiao2018simple} are adopted to predict the body landmarks, and then based on which the part features are learned \cite{su2017pose,zhao2017spindle,kalayeh2018human,sarfraz2018pose,suh2018part,xu2018attention}.
Compared to the extra task for pose estimation, recently more works are directly and uniformly part the feature maps to learn the body features \cite{zhang2017alignedreid,sun2018beyond,wang2018learning,fu2018horizontal,liu20183}.
Part-based convolutional baseline (PCB) \cite{sun2018beyond} adopts the uniform partition strategy with identity supervision on every local part, and the refined part pooling method to enhance the part representations.
Wang et.al \cite{wang2018learning} carefully designed a multi-branch deep network architecture, multiple granularity network (MGN), consisting of one branch for global feature representations and two branches for local feature representations, to learn features with various granularities.
Moreover, horizontal pyramid matching (HPM) \cite{fu2018horizontal} was proposed to use partial feature representations at different horizontal pyramid scales, to enhance the discriminative capabilities of various person parts.
CA$^{3}$Net \cite{liu20183} was proposed to learn the appearance features from both horizontal and vertical body parts of pedestrians with spatial dependencies among body parts, simultaneously exploiting the semantic attributes of person.

\textbf{Attention:} Alternatively, more and more researchers are adopting the attention mechanism to focus on the desired body parts.
Liu et.al \cite{liu2017end} proposed a soft attention based model, comparative attention network (CAN), to selectively focus on parts of pairs of person images with long short-term memory (LSTM) \cite{hochreiter1997long} method.
Li et.al \cite{li2017learning} designed a multi-scale context-aware network (MSCAN) to learn powerful features over full body and body parts, where the body parts are learned by the hard attention using the spatial transformer networks (STN) \cite{jaderberg2015spatial} with spatial constraints.
Zhao et.al \cite{zhao2017deeply} realised the attention models through a deep convolutional networks to compute the representations over part regions.
Liu et.al \cite{liu2017hydraplus} proposed a new attention-based model, hydraPlus-net (HP-net), which multi-directionally feeds the multi-level attention maps to different feature layers. Hp-net is capable of capturing multiple attentions from low-level to semantic-level, and explores the multi-scale selectiveness of attentive features to enrich the final feature representations for a pedestrian image.
Li et.al \cite{li2018harmonious} formulated a novel harmonious attention CNN (HA-CNN) model for joint learning of soft pixel attention and hard regional attention along with simultaneous optimisation of feature representations, dedicated to optimise person Re-ID in uncontrolled (misaligned) images.
Zheng et.al \cite{zheng2018re} proposed the consistent attentive siamese network, providing mechanisms to make attention and attention consistency end-to-end trainable in a siamese learning architecture. It is an effective technique for robust cross-view matching as well as can explain the reason why the model predicts the two images to belong to the same person.
Wang et.al \cite{wang2018mancs} proposed a multi-task attentional network with curriculum sampling (MANCS) method for Re-ID from the following aspects: fully utilizing the attention mechanism for the person misalignment problem and properly sampling for the ranking loss to obtain more stable person representation.

Different from those existing attention-based Re-ID methods, we introduce two type of attentions to adaptively extract discriminative person features for enhancing the model generalization and adaptation for cross-domain person Re-ID. We also systematically explore the possible arrangements of the two type of attentions, the combination of spatial and channel attentions, and hierarchical attention.

\begin{figure*}[t]
 \centering
 \includegraphics[width=180mm]{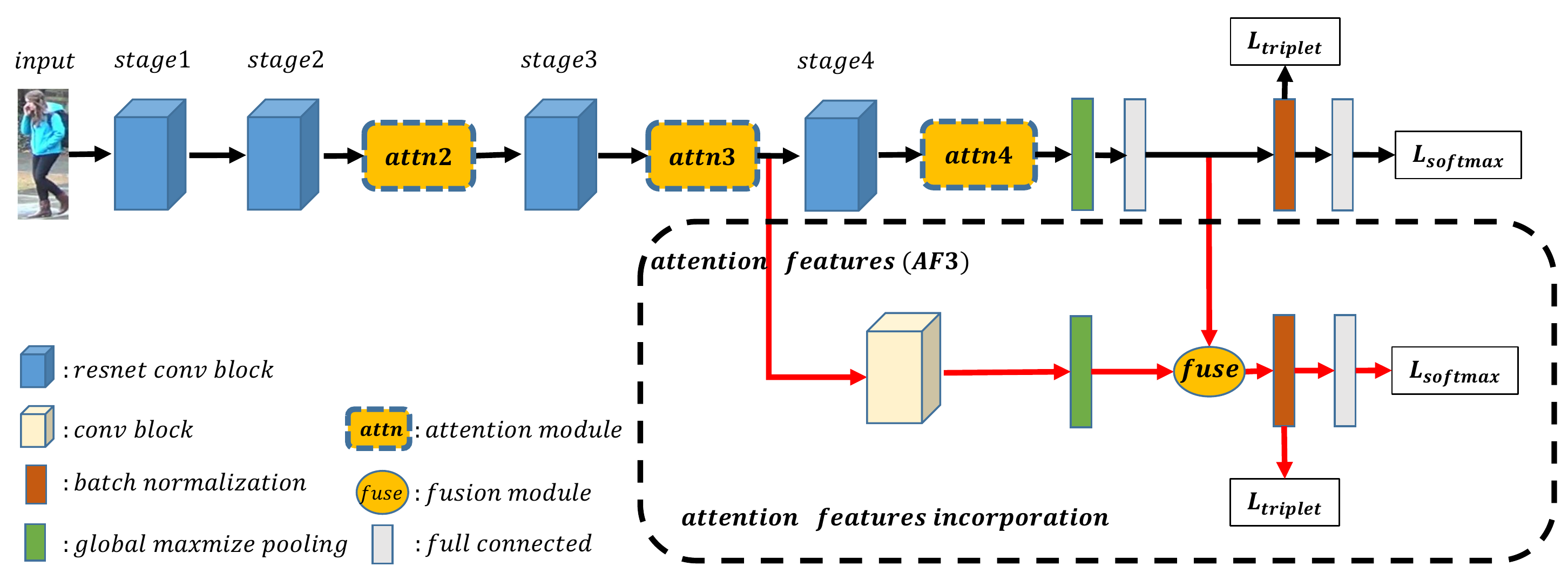}
 \caption{The framework of attention-based discriminative feature learning model. For example, taking the ResNet50 model as the backbone, we can simply incorporate the attention modules ($attn$) after each stage, extract the attention features, and then fuse them with the backbone output to obtain the final person features. The black lines without attention modules are the baseline procedures, while adding those red lines are procedures of skip connections for additional attention features incorporation.}
 \label{fig:framework}
 \end{figure*}
 
\textbf{Cross-domain:} As to the cross-domain property of person Re-ID, it refers to high requirements of model generalization and adaptation. Recently, more and more researches are concentrating on incorporating the properties of target-domain into the model training with the source-domain data.
Wang et.al \cite{wang2018transferable} introduced a transferable joint attribute-identity deep learning (TJ-AIDL) model for simultaneously learning an attribute-semantic and identity discriminative feature representation in a multi-task learning framework, which can be transferrable to any unseen target domain.
With the big success of CycleGAN \cite{zhu2017CycleGAN} in image translation, similarity preserving generative adversarial network (SPGAN) \cite{deng2018image} and person transfer generative adversarial network (PTGAN) \cite{wei2018person}, were proposed to transfer the source-domain images into the target-domain style, then train those translated images with their corresponding labels in the source domain.
Lv et.al \cite{lv2018unsupervised} proposed an unsupervised incremental learning algorithm, TFusion, to use the abundant unlabeled data in the target domain by adding the transfer learning of the pedestrians' spatio-temporal patterns in the target domain.
Huang et.al \cite{huang2018eanet} proposed the enhancing alignment networks (EANet) to address the cross-domain Re-ID task. EANet constits of two new modules: part aligned pooling (PAP) and part segmentation (PS), both of which are based on the body keypoints predicted by the pose estimation model.

All of the aforementioned cross-domain methods would leverage those unlabeled target-domain data. We are here trying to directly exploit our model training only on the source-domain data to perform the cross-domain Re-ID task, through a simple way to incorporate the attention modules.

\section{The framework of attention-based discriminative feature learning}
\label{sec:framework}
The goal of this paper is to perform the cross-domain person Re-ID under the single dataset training setting. The most important point is to enhance the generalization and adaptation of model trained only with the source domain data. \textbf{We argue that those models essentially with the ability of adaptively extracting the discriminative features on different datasets are with high generalization and adaptation.} Therefore, how to adaptively extract the discriminative features on different datasets is the key point to focus on. To address this problem, we propose to introduce the attention mechanism for discriminative feature learning. In summary, we incorporate attention mechanism to enhance the model generalization and adaptation for cross-domain person Re-ID.

In this section, we will introduce our simple framework of attention-based discriminative feature learning (ADFL).
Based on the popular attention modules (Sec. \ref{sec:attn}), we try to use a simple way to incorporate them into the well designed deep convolutional neural networks, such as ResNet50, in order to boost the cross-domain person Re-ID performance. 

Fig. \ref{fig:framework} depicts the detailed structure of our proposed ADFL model, which mainly consists of three components: the backbone, the attention modules incorporation and the skip connections for additionally incorporating the attention features.

\subsection{Backbone network}
\label{ssec:network}
ADFL can take any deep neural network designed for image classification as the backbone, e.g., Google Inception \cite{szegedy2017inception} and ResNet \cite{he2016deep}. We would take the ResNet50 model as an example to perform the Re-ID task, with the consideration of its competitive performance in some Re-ID systems \cite{sun2018beyond,wang2018mancs} as well as its relatively concise architecture.
ResNet50 model mainly consists of four res-convolution blocks, $stage1$, $stage2$, $stage3$ and $stage4$, as illustrated in Fig. \ref{fig:framework}.
The modifications to the original ResNet50 model are as following.
\begin{enumerate}
\item We employ no down-sampling operations in the $stage4$ block to preserve more areas of reception fields for local (body part) features.
\item The global max pooling layer substitutes for the global average pooling layer after the $stage4$.
\item Then a full connected layer is added as a bottleneck to reduce the dimensions if necessary.
\item A batch normalization ($BN$) layer is added sequentially, whose output would be adopted to perform the metric learning with triplet loss \cite{hermans2017defense}.
\item Finally a full connected layer with desired dimensions (corresponding to the number of identities of person in our model) is adopted to perform the classification with softmax loss.
\end{enumerate}

The procedure of backbone network follows the black lines without the attention modules, as illustrated in Fig. \ref{fig:framework}.

\subsection{Attention modules incorporation}
\label{ssec:atten_incor}
Since our attention modules (Sec. \ref{sec:attn}) are all in a size-identical map manner, that the output and input are with identical size, we can incorporate our attention modules into ResNet50 model at any position.
For simplicity, as illustrated in Fig. \ref{fig:framework} we only consider the following 3 simple cases:
\begin{enumerate}
\item Incorporating the attention module after $stage2$. We term it as $attn2$.
\item Incorporating the attention module after $stage3$. We term it as $attn3$.
\item Incorporating the attention module after $stage4$. We term it as $attn4$.
\end{enumerate}

\subsection{Additional attention features incorporation}
\label{ssec:mlf_incor}
Attention modules can dynamically focuses on salient parts according to the detailed content of each image, which can make a information integration to emphasise the salient aspects.
Moreover, those attention outputs are just the mid-level features in terms of the model outputs.
To directly take advantage of those attention features, we try to fuse them with those features before the final $BN$ layer of backbone network to obtain the final person features. Therefore,  the final person features are with both high and middle level semantic visual information.

As illustrated in Fig. \ref{fig:framework}, taking the output of $attn3$ for example, the red lines depicts the procedures for incorporating additional attention features (AF).

\begin{figure}
\centering
\includegraphics[width=80mm]{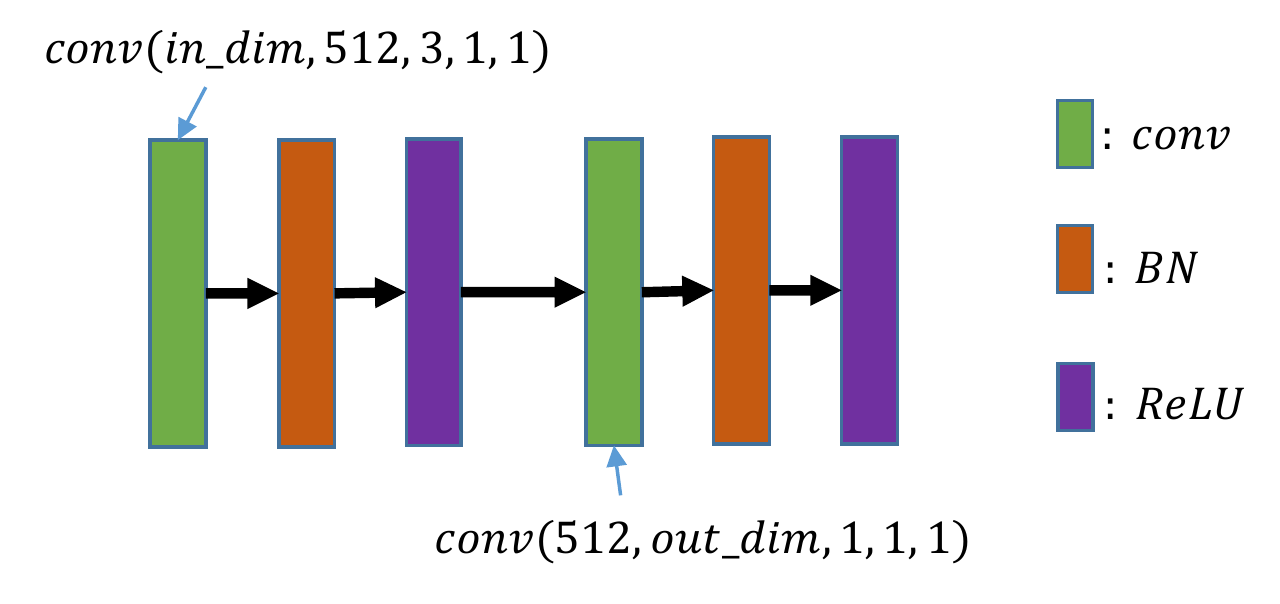}
\caption{The illustration of the convolutional block. One $3 \times 3$ convolutional layer is followed by a batch normalization layer and a ReLU layer; then another $1 \times 1$ convolutional layer followed with a batch normalization layer and a ReLU layer is stacked. The parameters ($in\_dim$ and $out\_dim$) of the two convolutional layers can be changed according to the input and output from the backbone network.}
\label{fig:conv}
\end{figure}

\begin{figure*}
  \centering
  \begin{tabular}{c@{\hspace{5mm}}c}
  \includegraphics[width=90mm]{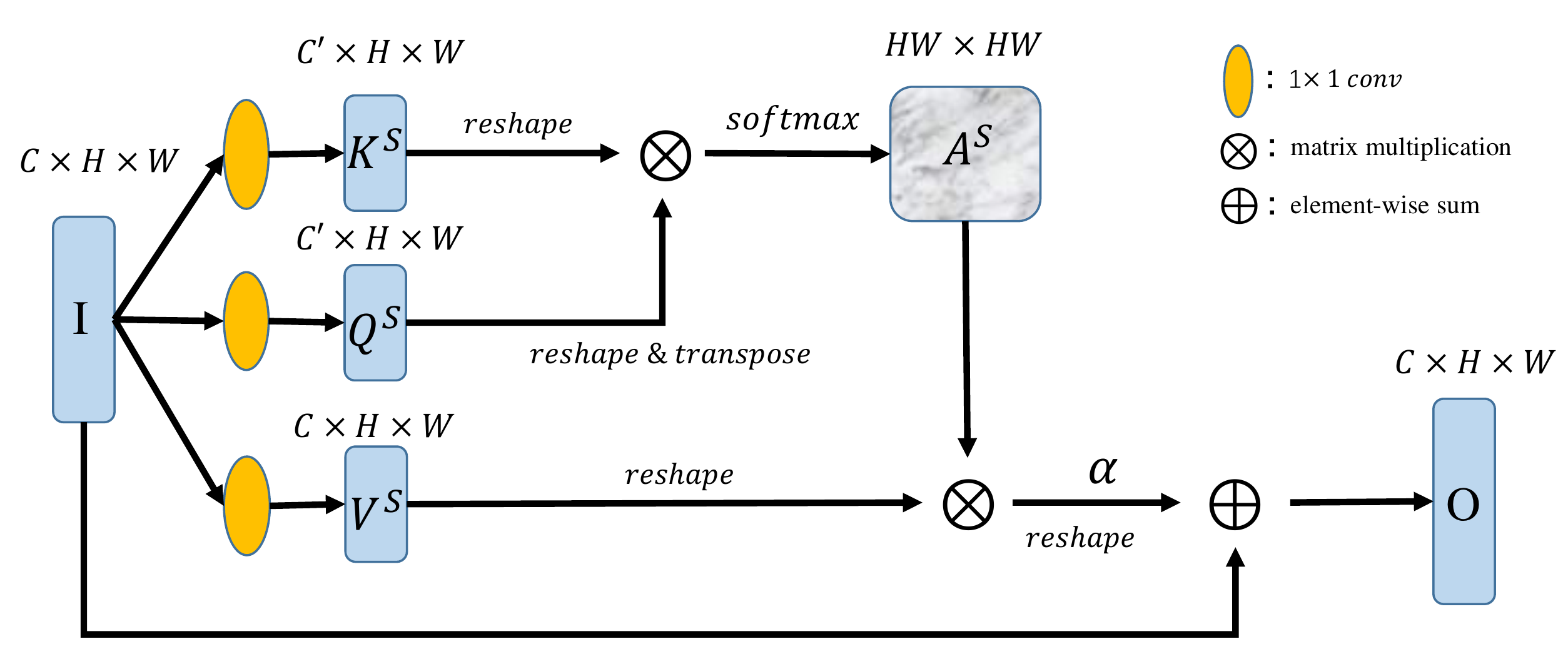} &
  \includegraphics[width=90mm]{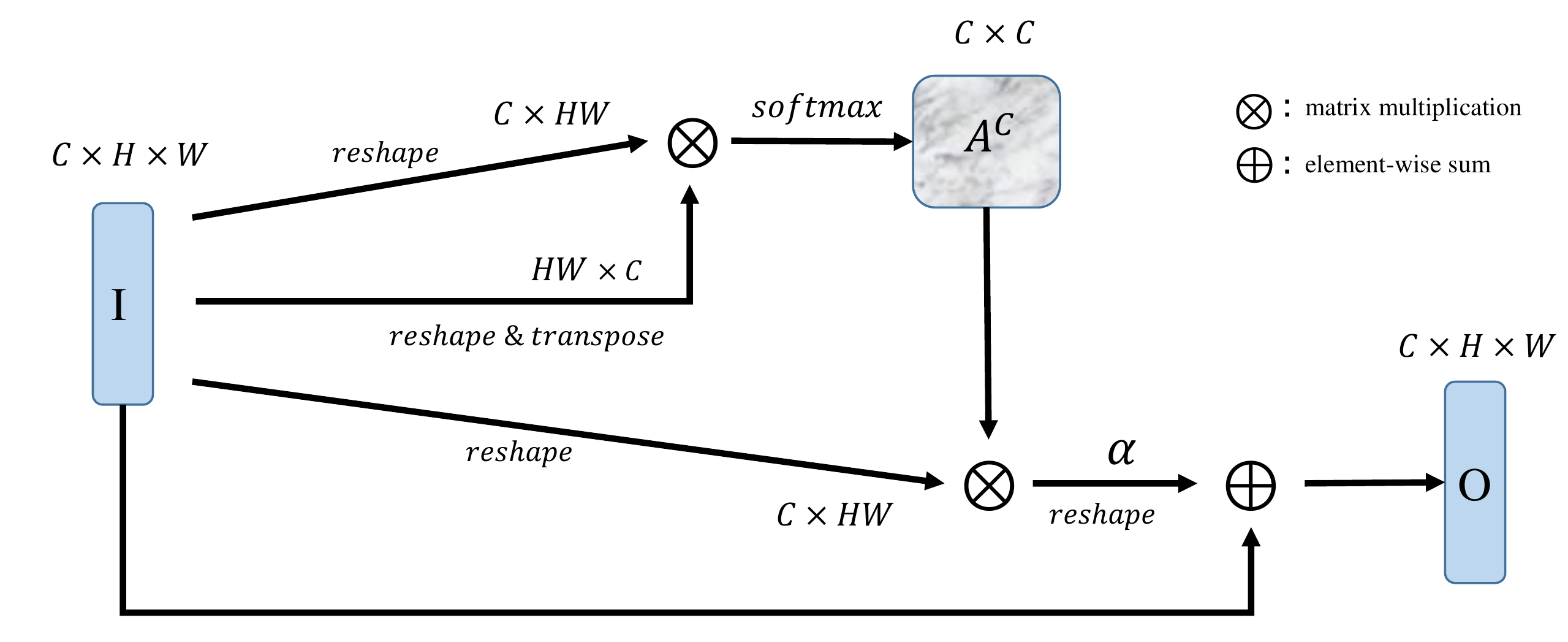}\\
  (a) $1s$ &(b) $1c$ \\
  \includegraphics[width=90mm]{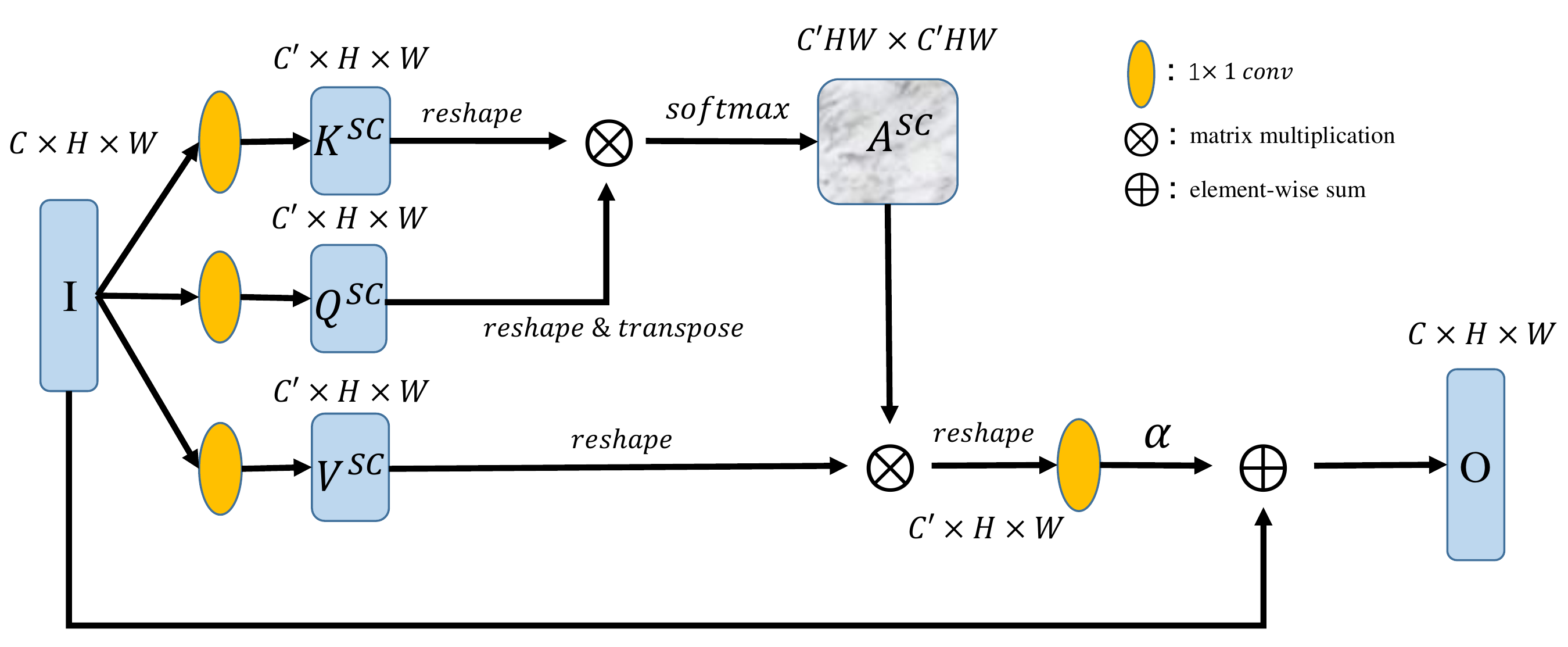} &
  \includegraphics[width=90mm]{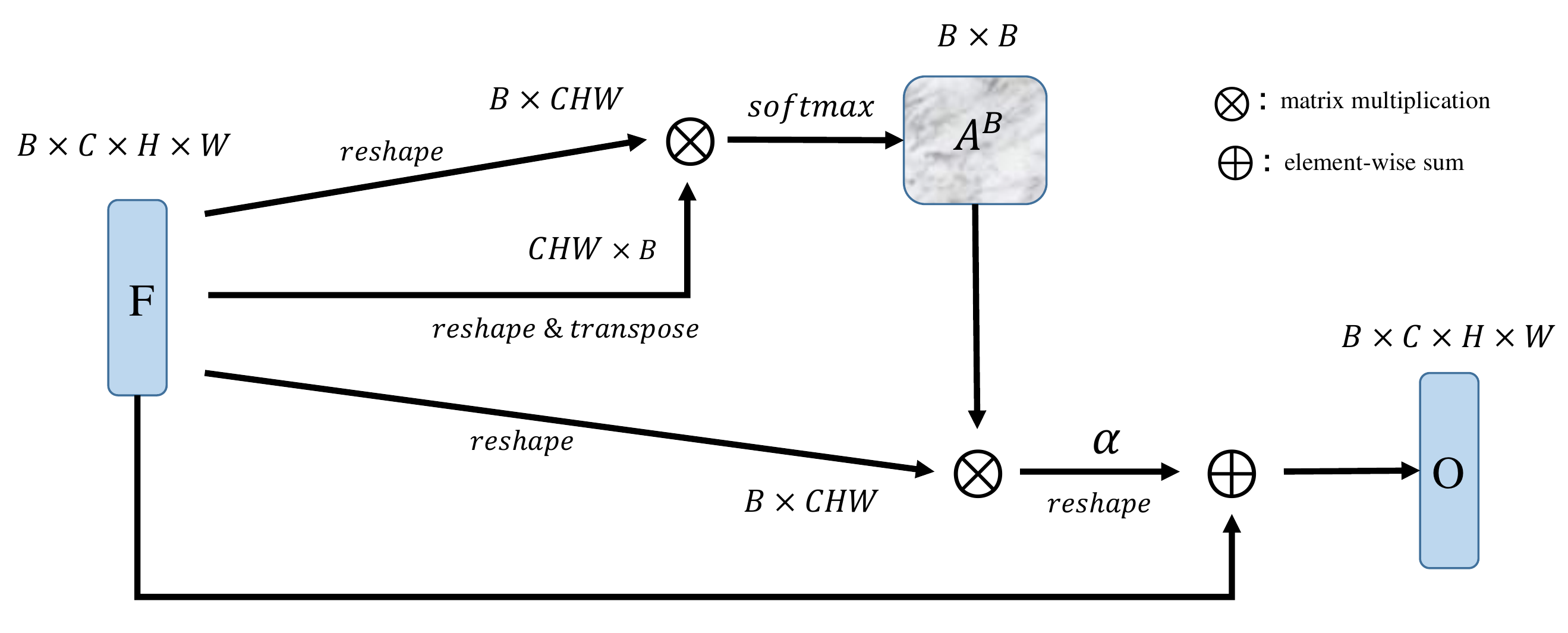}\\
  (c) $1h$ &(d) $1b$
  \end{tabular}
  \caption{Type I: long-range dependency based attention. The details of the attention modules are respectively illustrated in (a) $1s$, the spatial attention , (b) $1c$, the channel attention, (c) $1h$, the hyper attention combining the spatial and channel attention, and (d) $1b$, the batch attention.}
  \label{fig:atten1}
 \end{figure*}
 
\begin{enumerate}
\item The attention features should pass a convolutional block to change the feature map size according to the feature map size of backbone network. We design the convolutional block in a bottleneck manner consisting of two convolutional layers, each of which is followed with a batch normalization layer and a ReLU layer, as illustrated in Fig. \ref{fig:conv}.
\item A global max pooling layer is followed, similar to the backbone network.
\item The attention features and the backbone features are fused to obtain the final person features. For simplicity, we only consider the fusion mechanism to be sum or concatenation, which would be explored in the experiments.
\item Similar to the backbone network, a batch normalization layer with triplet loss and then a full connected layer with softmax loss are adopted.
\end{enumerate}

\section{Attention modules}
\label{sec:attn}
We argue that the attention mechanism is inherent with the characteristic of discriminative feature learning.
The attention mechanism plays an important role in human visual perception system, where human visual perception does not process the whole image at once, but exploits a sequence of partial glimpses and selectively focuses on salient parts to well capture the visual information.

For the person Re-ID task, as discussed in Sec. \ref{sec:intro}, the partial body features play an important role to distinguish one person from others. The partial body features are intuitively learned through the spatial partition from the human visual perception view, which corresponds to the \textbf{spatial attention}. Spatial attention mainly focuses on ``where'' are the informative parts given an person image.

What's more, in the community of deep convolutional neural networks, due to the adoption of fully connected layer and the softmax loss, each channel map of high level feature maps can be regarded as a class-specific response, and different semantic responses are associated with each other \cite{zeiler2014visualizing}. In person Re-ID task, different sematic classes maybe correspond to the salient body parts. It indicates that different person body parts cause different responses corresponding to different channels in the feature map. Therefore, by exploring the interdependencies between channel maps, which corresponds to the \textbf{channel attention}, we could improve the person features with specific semantics for body parts. Channel attention mainly focuses on ``what'' are the most meaningful parts given an person image.

In this section, we apply two type of attention mechanisms from different aspects to well extract the discriminative person features. One type is to capture the long-range dependency of contextual information, while the other is to directly generate the attention maps. Both of the two type of attention mechanisms take the spatial dimension and channel dimension into account. Finally, different aggregations of the spatial attention and channel attention for further refinement will be briefly described.

Note that the attention mechanisms are well explored in many literatures \cite{vaswani2017attention,wang2018non,zhao2017deeply,liu2017hydraplus,li2018harmonious,zheng2018re,wang2018mancs}. Only some subtle modifications are accordingly made. Therefore, we only outline those attention modules in Figs. \ref{fig:atten1} and \ref{fig:atten2}. The computation details of attention modules can be find in our supplementary materials.

\subsection{Type I: long-range dependency based attention}
\label{ssec:lr_attn}
The long-range dependency is emphasized in transformer \cite{vaswani2017attention} and non-local network \cite{wang2018non}, which computes the response at a position as a weighted sum of features at all positions in the input feature maps.
\begin{align}
    O_j = \sum_{i=1}^{N} A_{j,i} I_j,  \label{eq:1attn}
\end{align}
where $I$ is the input with $N$ positions, $A_{j,i}$ is the attention map measuring the $i^{th}$ position's impact on the $j^{th}$ position, $O$ is the output with the same size to $I$\footnote{Note that all of the attention modules in this paper are in a size-identical map manner, that the output and input are with identical size.}.

Here, we exploit the effectiveness of self-attention mechanism from the long-range dependency relationship aspect to represent the person features.
Given an intermediate feature map $F \in R^{B\times C \times H \times W}$ as input, where $B$, $C$, $H$ and $W$ are batch size, channel, height and width, we design the attention modules for spatial, channel, spatial+channel (hyper) and batch dimensions, respectively. All of the attention modules are shown in Fig. \ref{fig:atten1}. The detailed calculation procedures of these attention modules can be found in supplementary materials.

\subsection{Type II: direct generation based attention}
\label{ssec:dg_attn}

\begin{figure}
  \centering
  \begin{tabular}{c}
  \includegraphics[width=90mm]{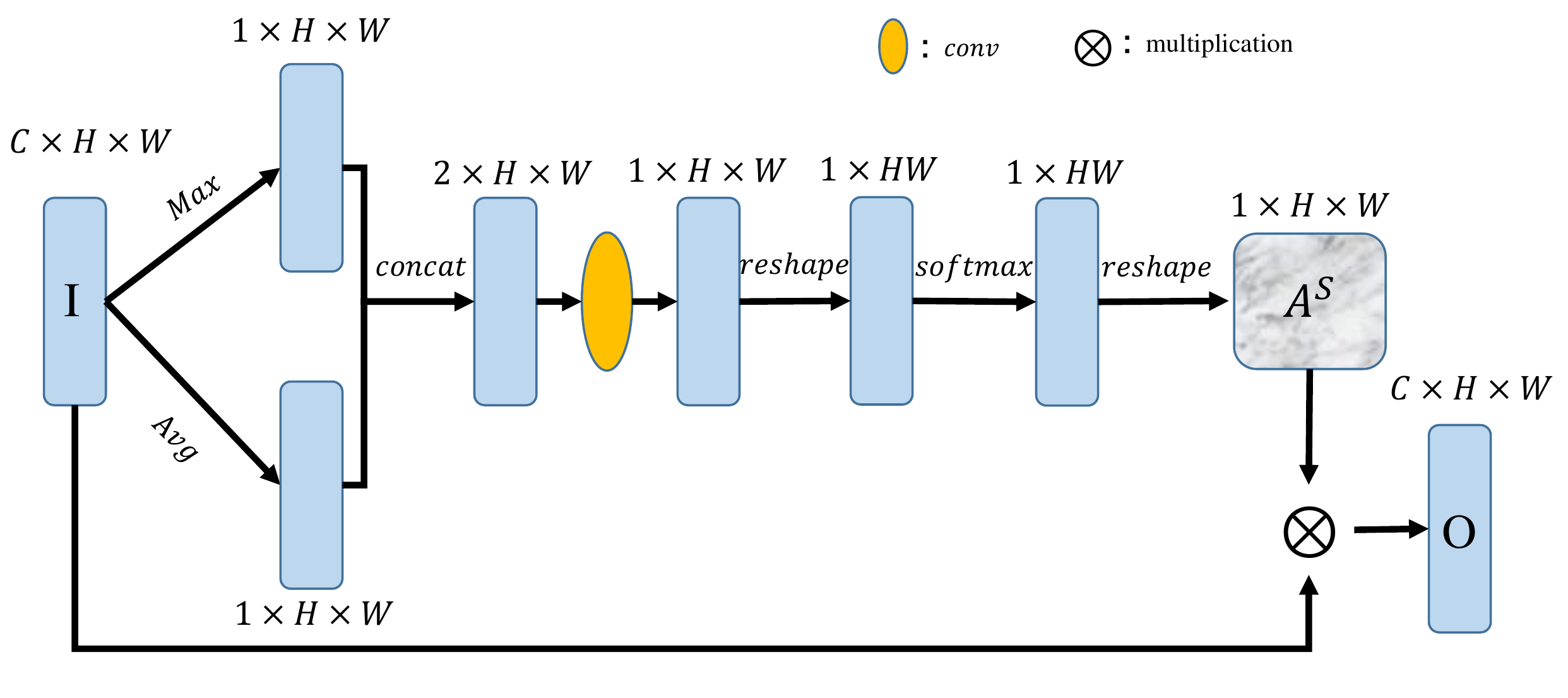}\\   (a) $2s$ \\
  \includegraphics[width=90mm]{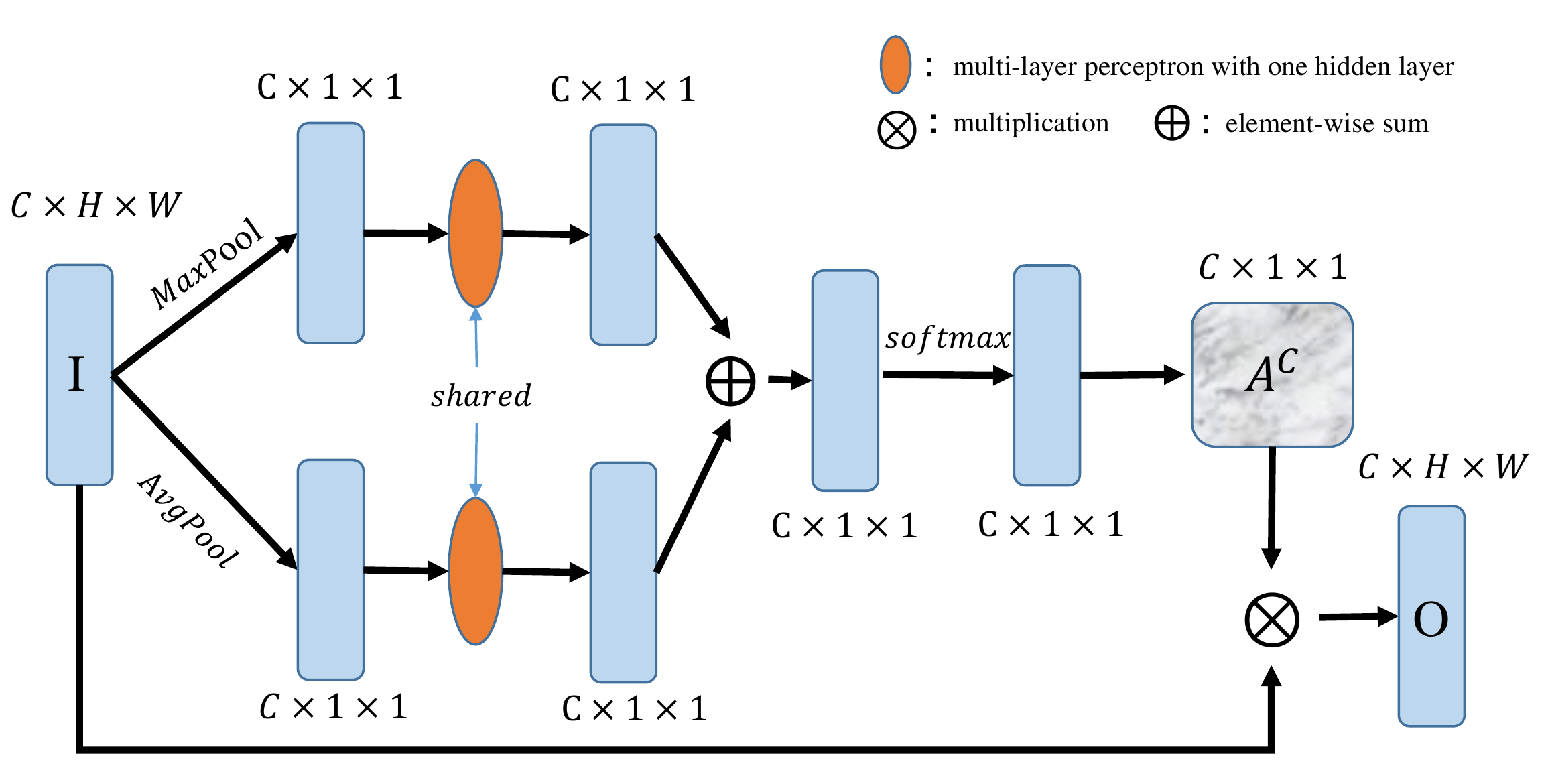}\\   (b) $2c$ \\
  \includegraphics[width=90mm]{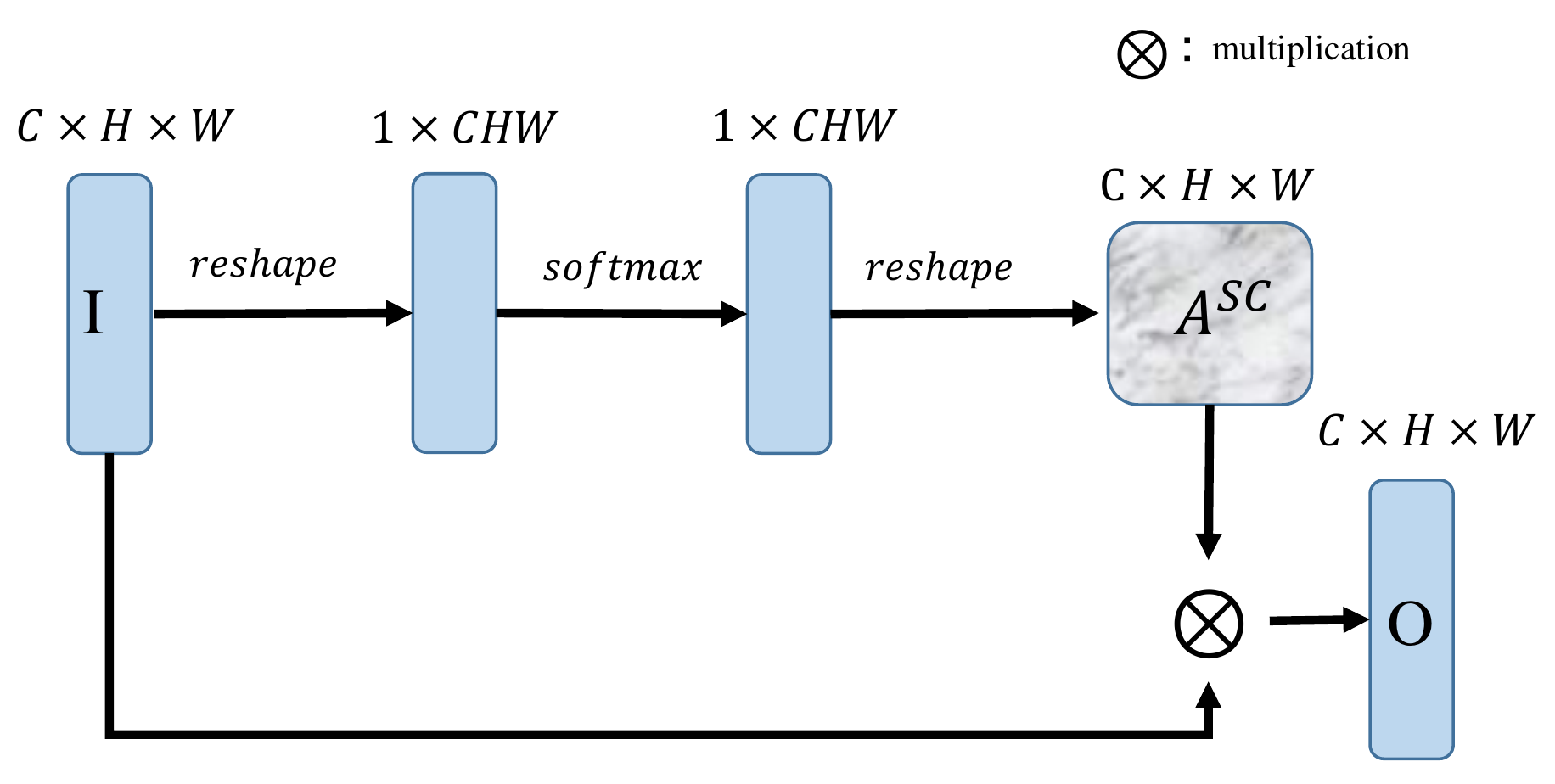}\\  (c) $2h$
  \end{tabular}
  \caption{Type II: direct generation based attention. The details of the attention modules are respectively illustrated in (a) $2s$, the spatial attention, (b) $2c$, the channel attention, and (c) $2h$, the hyper attention combining the spatial and channel attention.}
  \label{fig:atten2}
 \end{figure}

Another type of attention is intuitive to assign different weights for different positions.
\begin{align}
    O_i = A_i I_i, \label{eq:2attn}
\end{align}
where $I$ is the input feature map, $A_i$ is the attention weight for the $i^{th}$ position, and $O$ is the attention output.

We design the direct generation based attention for spatial, channel and spatial+channel (hyper), respectively. All of the attention modules are shown in Fig. \ref{fig:atten2}. The detailed calculation procedures of these attention modules also can be found in supplementary materials. Compared to the existing direct generation based attention works \cite{li2018harmonious}, we specially add the $softmax$ operation over the attention map to emphasize the importance of one position related to all the other positions, following the philosophy of long-range dependency in the last section (Sec \ref{ssec:lr_attn}).

\subsection{Arrangements of the spatial and channel attentions}
\label{ssec:arrang}
\begin{figure}
  \centering
  \begin{tabular}{c|c}
  \includegraphics[width=40mm]{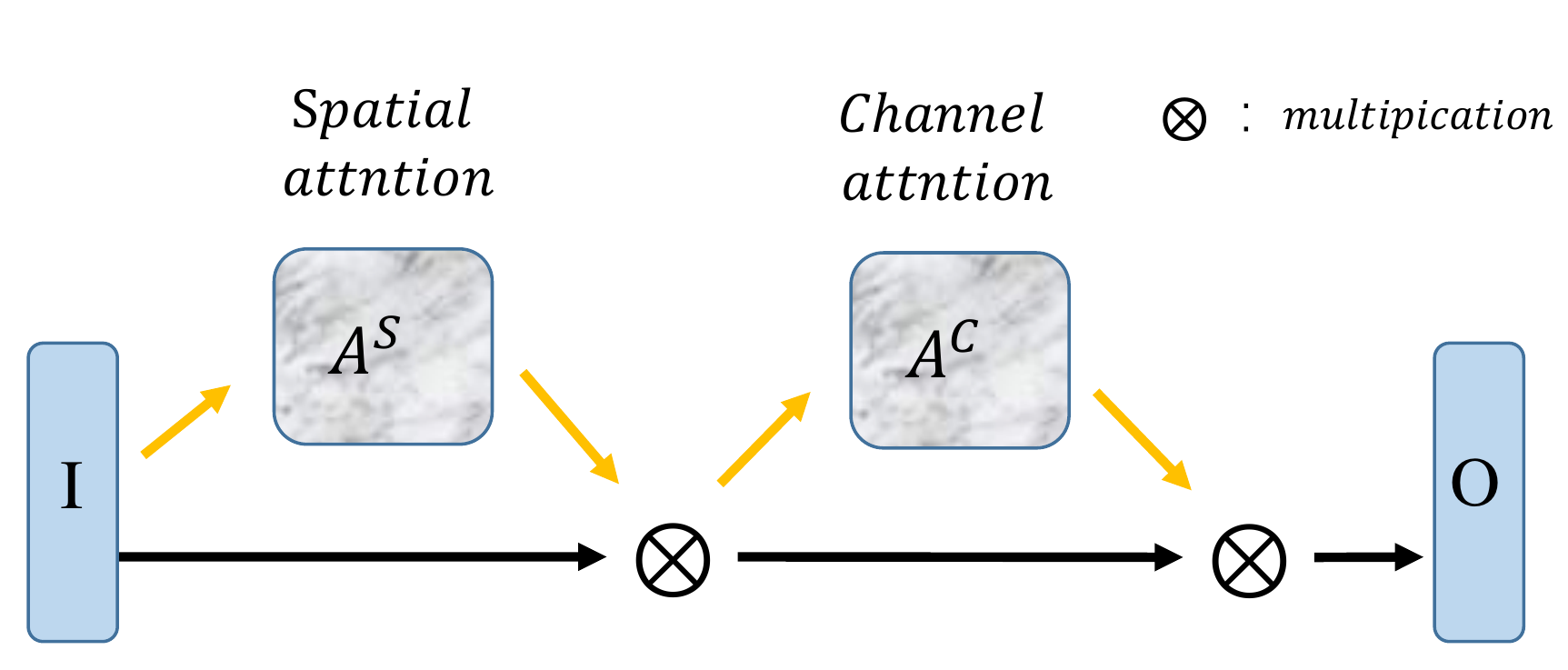}&
  \includegraphics[width=45mm]{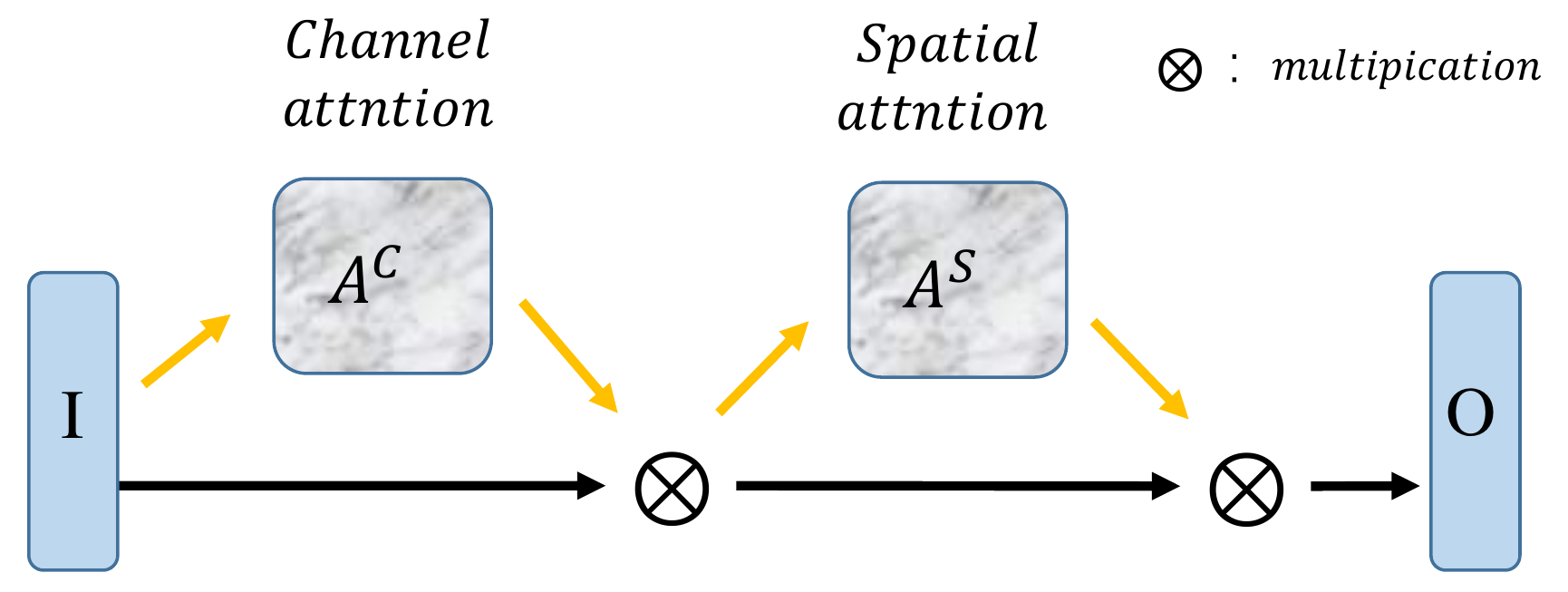}\\
  (a) $sc$ & (b) $cs$\\
  \includegraphics[width=40mm]{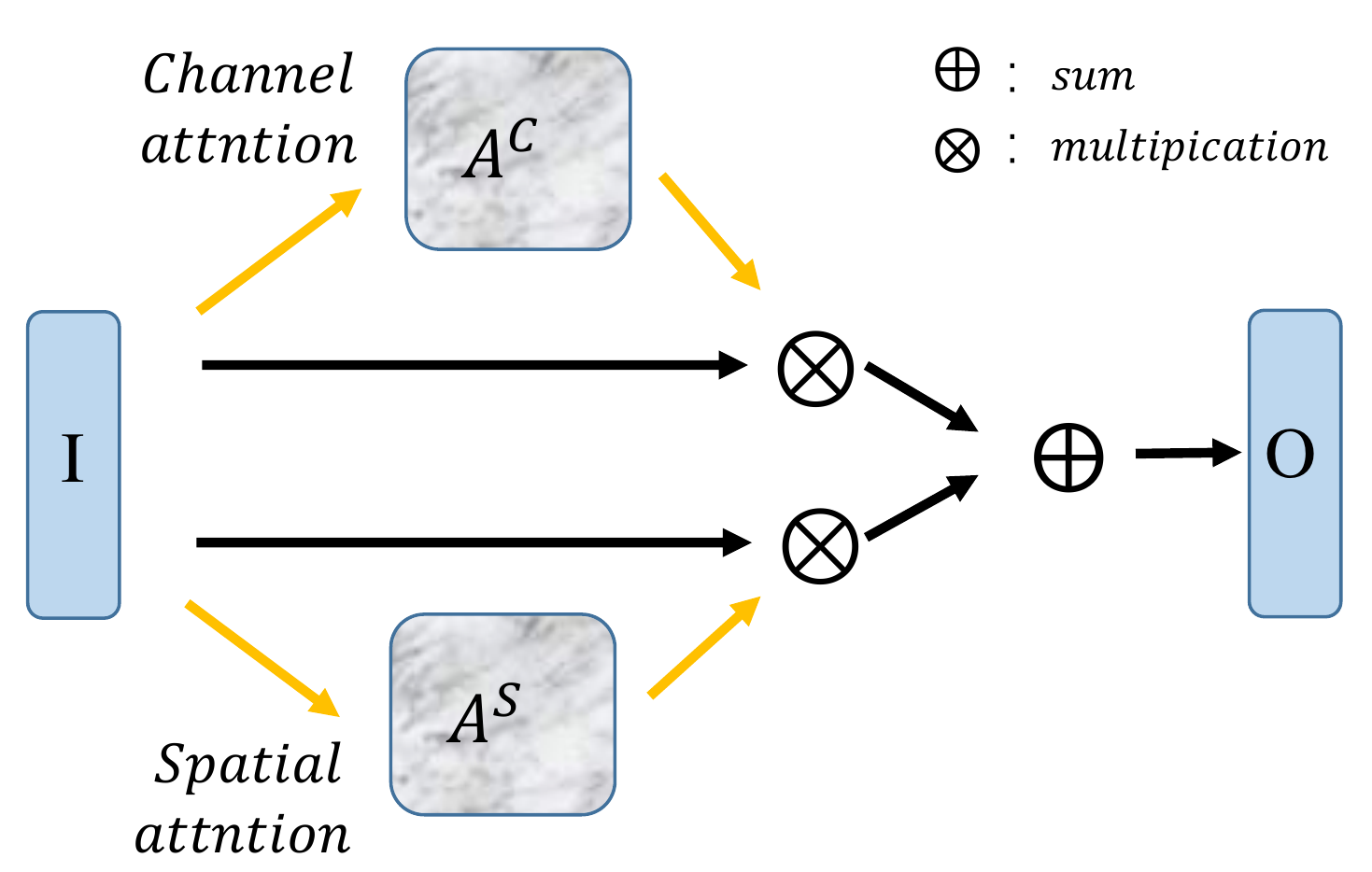}&
  \includegraphics[width=45mm]{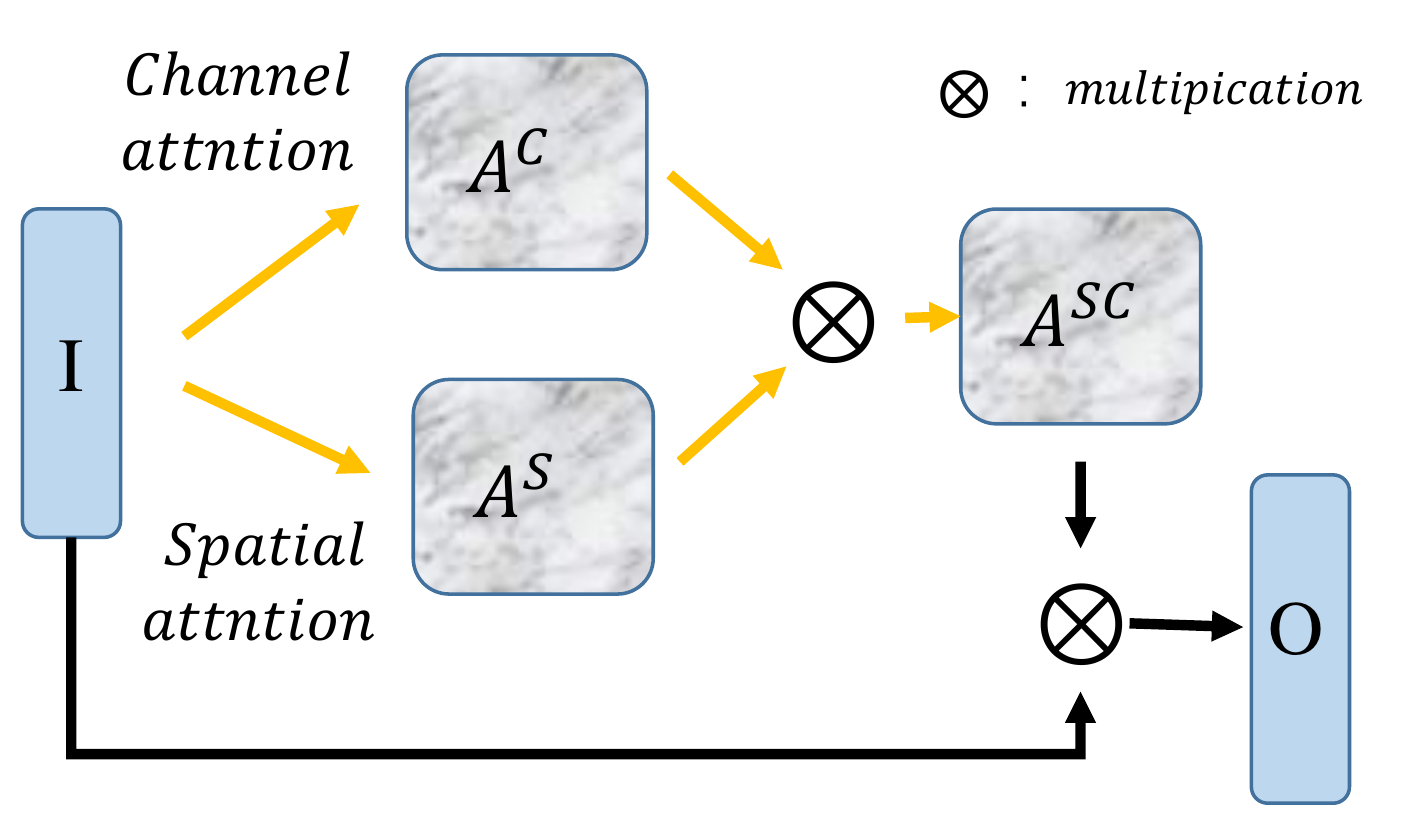}\\
  (c) $sum$ & (d) $multiply$
  \end{tabular}
  \caption{The arrangements of the spatial and channel attentions. The sequential arrangements: (a) $sc$, spatial attention first, then followed by channel attention ; (b) $cs$, channel attention first, then followed by spatial attention. The parallel arrangement: (c) $sum$, spatial attention and channel attention are performed simultaneously and then summed together. Especially for type II attention, the multiplication arrangement: (d) $multiply$, the spatial and channel attention map are multiplied together to generate the whole spatial-channel attention map, and then multiply the input. }
  \label{fig:arrange}
 \end{figure}

 Given an input image, the hyper attentions (in Fig. \ref{fig:atten1} (c) and Fig. \ref{fig:atten2} (c)) model the spatial and channel attentions simultaneously. However, when spatial and channel attentions are separately calculated, we can arrange them in a sequential or parallel manner for both the two types of attention.

 As illustrated in Fig. \ref{fig:arrange} (a) and (b), the spatial and channel attentions can be sequentially arranged to obtain the final attention results. Or the spatial attention and channel attention could be performed simultaneously in a parallel manner, and then summed together to obtain the final attention results, as shown in Fig. \ref{fig:arrange} (c).

 For the type II attention (direct generation based attention, Sec. \ref{ssec:dg_attn}), which can calculate the attention maps along the spatial and channel dimension, respectively, obtaining a 2D spatial attention map $A^{S} \in R^{1 \times H \times W}$ and a 1D channel attention map $A^{C} \in R^{C \times 1 \times 1}$, we consider to perform the multiplication to obtain a 3D spatial-channel attention map $A^{SC} \in R^{C \times H \times W}$, as illustrated in Fig. \ref{fig:arrange} (d).
 \begin{align}
    A^{SC} = A^{C} \otimes A^{S},
 \end{align}
 where $\otimes$ is the multiplication with broadcast mechanism accordingly.

\section{Experiments and analysis}
\label{sec:exp}
In this section, we evaluate the effectiveness of our proposed ADFL model for person Re-ID tasks with three datasets, Market-1501 \cite{zheng2015scalable}, DukeMTMC-reID \cite{ristani2016performance}, and MSMT17 \cite{wei2018person}, under both of the single-domain and cross-domain settings.

\subsection{Datasets and protocols}
\label{ssec:datasets}
This section will briefly introduce these datasets and the evaluation protocols.

\textbf{Market-1501.} This dataset contains 32,668 annotated bounding boxes, predicted by DPM detector \cite{felzenszwalb2008discriminatively}, of 1,501 identities which are taken from 6 different cameras. The identities are split into 751 training IDs with 12,936 images and 750 query IDs. There are in total 3,368 query images, each of which is randomly selected from each camera so that cross-camera search can be performed. The retrieval for each query image is conducted over a gallery of 19,732 images including 6,796 junk images.

\textbf{DukeMTMC-reID.} This dataset consists of 36,411 images of 1,812 identities from 8 high-resolution cameras. Among them, 16,522 images of 702 identities are randomly selected from the dataset as the training set. While the testing set consists of the remaining 1,110 identities, among which 2,228 images of 702 identities are as query, and 17,661 images of the 1,110 identities are as gallery set.

\textbf{MSMT17.} It is a large and challenging Re-ID dataset with 126,441 annotation bounding boxes of 4,101 identities from 15 cameras, involving wide light varieties and different weather conditions. The bounding boxes are predicted by Faster-RCNN \cite{ren2015faster}. The training set contains 32,621 bounding boxes of 1,041 identities, and the testing set contains 93,820 bounding boxes of 3,060 identities. From the testing set, 11,659 bounding boxes are randomly selected as query images and the other 82,161 images are used as gallery images.

\textbf{Protocols.} We adopt the cumulative matching characteristics (CMC) at rank-1 and the mean average precision (mAP) as the evaluation indicators to report the performances of different Re-ID methods on these Re-ID datasets.

\subsection{Implementation details}
\label{ssec:implement}
The implementation of our proposed method is based on the Pytorch framework.
\textbf{Model.} We adopt the  ResNet50 model pre-trained on ImageNet as the backbone network of our proposed model, changing the stride of convolution in $stage4$ from 2 to 1.
\textbf{Preprocessing.} To capture more detailed information from the person images, in training phase the input image is resized to $384 \times 128$, and padded with 10. Then the input image is randomly left-right flipped and cropped to $384 \times 128$ for data augmentation. Left-right image flipping is also utilized in the testing phase.
\textbf{Mini-batch size.} Each mini-batch is sampled with randomly selected $P$ identities and randomly sampled $K$ images for each identity from the training set to cooperate the requirement of triplet loss. Here we set $P = 8$ and $K = 4$, leading to a mini-batch size of 32 to train our proposed model.
\textbf{Optimization.} We use the adam method \cite{kingma2014adam} as the optimizer with $\beta_1 = 0.9$ and $\beta_2 = 0.999$. The network is trained for 150 epochs in total. Those layers in backbone network are fixed for 5 epochs firstly.
\textbf{Learning rate.} (1) The initial learning rate is set as 3.5e-4 and weight decay is set as 5e-4. (2) Increasing the learning rate in the early 20 epochs from 3e-6 to 3.5e-4 with a linear warmup strategy. (3) The learning rate is decayed with a parameter 0.1 at epoch 80 and 130, respectively. Overall, the learning rate ($lr$) is adjusted as follows,
\begin{align}
    lr(t) =
    \left\{
    \begin{tabular}{cl}
        $3 \times 10^{-6} + \frac{3.5 \times 10^{-4} - 3 \times 10^{-6}}{20} t$, & $t \leq 20$ \\
        $3.5 \times 10^{-4}$, & $20 < t \leq 80$ \\
        $3.5 \times 10^{-5}$, & $80 < t \leq 130$ \\
        $3.5 \times 10^{-6}$, & $130 < t \leq 150$
    \end{tabular}.
    \right.
\end{align}

\begin{table}
\caption{Evaluating the effectiveness of our strong baseline compared to those without the utilized components and some recently published state-of-the-art results by listing the Rank-1 and mAP results (\%). Label M and D denote Market-1501 to DukeMTMC-reID, respectively. M$\to$D means training on M and testing on D.}
\label{tab:baseline}
  \centering
  \begin{tabular}{l|c|c|c|c}
    \toprule[2pt]
    \multirow{2}{*}{Method}  & \multicolumn{2}{c|}{M$\to$M} & \multicolumn{2}{c}{M$\to$D} \\
     \cline{2-5}
                      & Rank-1    & mAP   & Rank-1 & mAP    \\ \toprule[1pt]
     $256 \times 128$ & 93.7 & 82.7  & 48.6 &  30.2  \\
     only softmax & 93.5 & 81.4 & 48.7 & 29.1   \\
     only triplet & 90.4 & 76.2 & 39.1 &  22.1  \\
     not fix backbone& 93.3 & 81.5 & 45.1 & 27.0   \\
     no warmup & 93.5 & 82.4 & 47.8 &  29.0  \\
     no $BN$ & 93.5 & 80.9 & 48.8 & 30.4   \\
     baseline (ours) & \textbf{94.0} & \textbf{83.1} & \textbf{48.9} & \textbf{30.8}   \\ \toprule[1pt] \toprule[1pt]
      & \multicolumn{2}{c|}{M$\to$M} & \multicolumn{2}{c}{Publication} \\ \hline
     IDE \cite{zheng2016person} & 73.9 & 47.8 & \multicolumn{2}{c}{arXiv16}  \\
     OIM \cite{xiao2017joint} & 82.1 &-  & \multicolumn{2}{c}{CVPR17}  \\
     SSM \cite{bai2017scalable} & 82.2 & 68.8 & \multicolumn{2}{c}{CVPR17}  \\
     SVDNet \cite{sun2017svdnet} & 82.3 & 62.1 & \multicolumn{2}{c}{ICCV17}  \\
     DaRe \cite{wang2018resource} & 86.4 & 69.3 & \multicolumn{2}{c}{CVPR18}  \\
     CamStyle \cite{zhong2018camera} & 89.5 & 71.6 & \multicolumn{2}{c}{CVPR18}  \\
     MLFN \cite{chang2018multi} & 90.0 & 74.3 & \multicolumn{2}{c}{CVPR18}  \\
     HACNN \cite{li2018harmonious} & 91.2 & 75.7 & \multicolumn{2}{c}{CVPR18}  \\
     E-CNN \cite{xiong2018towards} & 91.7 & 78.8 & \multicolumn{2}{c}{arXiv18} \\
     PCB \cite{sun2018beyond} & 92.3 & 77.4 & \multicolumn{2}{c}{ECCV18} \\
     PCB+RPP \cite{sun2018beyond} & 93.8 & 81.6 & \multicolumn{2}{c}{ECCV18} \\
     MANCS \cite{wang2018mancs} & 93.1 & 82.3 & \multicolumn{2}{c}{ECCV18} \\
    \toprule[2pt]
  \end{tabular}
\end{table}
\textbf{Features.} For the baseline and attention incorporation methods, the features after the final $BN$ layer of backbone network are adopted as the person representations during testing. While for the additional attention features incorporation method, the features after the other final $BN$ layer in the skip-connection flows (the red lines in Fig. \ref{fig:framework}) are adopted as the person representations during testing.

\subsection{The strong baseline}
\label{ssec:ex_baseline}

\begin{table*}
\caption{Verifying the effectiveness of those attention modules described in Sec. \ref{sec:attn} by listing the Rank-1 and mAP results (\%). Two types of attention are incorporated into the $stage2$ (corresponding to $attn2$), $stage3$ (corresponding to $attn3$) and $stage4$ (corresponding to $attn4$) of ResNet50 model, respectively. $s$: spatial attention, $c$: channel attention, $b$: batch attention, $h$: hyper attention, $cs$: channel attention first, then followed by spatial attention, $sc$: spatial attention first, then followed by channel attention, $sum$: spatial attention and channel attention are performed simultaneously and then summed together, $multiply$: the spatial and channel attention maps are multiplied together to generate the whole spatial-channel attention map.}
\label{tab:attn}
  \centering
  \begin{tabular}{l|c|c|c|c|c|c|c|c}
    \toprule[2pt]
    \multirow{4}{*}{Method}  & \multicolumn{8}{c}{Type I: long-range dependency based attention} \\
     \cline{2-9}
     & \multicolumn{4}{c|}{$attn2$} & \multicolumn{4}{c}{$attn3$}  \\ \cline{2-9}
     &\multicolumn{2}{c|}{M$\to$M} & \multicolumn{2}{c|}{M$\to$D}& \multicolumn{2}{c|}{M$\to$M} & \multicolumn{2}{c}{M$\to$D} \\ \cline{2-9}
                      & Rank-1    & mAP   & Rank-1 & mAP & Rank-1    & mAP   & Rank-1 & mAP    \\ \toprule[1pt]
      $1s$ &  \textbf{93.9}  &  81.8   & 49.3   & 31.4 & 93.9  &  \textbf{82.5}  & 52.1 & 33.9    \\
      $1c$ &  \textbf{93.9} &  81.8   &  \textbf{51.2}  & \textbf{32.3} & 93.8   &  \textbf{82.5}  & 53.8 & 34.8    \\
      $1b$ & 93.0$\pm$ 0.6  &  81.1$\pm$ 1.0   &  48.7$\pm$ 0.4  & 29.9$\pm$ 1.6 &  93.4$\pm$ 0.5   &  81.3$\pm$ 0.6  & 51.4$\pm$ 1.1 & 31.7$\pm$ 1.3 \\
      $1h$ & 93.5  &  81.9   &  49.3  & 31.7 & 93.7   &  82.1  & 53.5 & 33.7   \\
      $1cs$ & 93.7  &  \textbf{82.1}   &  50.3  & 32.0 & 93.6   & 82.2   & 53.9 & 35.2    \\
      $1sc$ & 93.8  &   82.0  &  49.9  & 31.7 & \textbf{94.1}   &  \textbf{82.5}  & 53.8 & 35.4   \\
      $1sum$ & 93.1  &  81.2   &  48.7  & 30.7 & 93.4   &  81.5  & \textbf{54.8} & \textbf{36.3}    \\
    \toprule[1pt]  \toprule[1pt]
    \multirow{4}{*}{Method}  & \multicolumn{8}{c}{ Type II: direct generation based attention} \\
    \cline{2-9}
    & \multicolumn{4}{c|}{$attn3$} & \multicolumn{4}{c}{$attn4$}  \\ \cline{2-9}
    &\multicolumn{2}{c|}{M$\to$M} & \multicolumn{2}{c|}{M$\to$D}& \multicolumn{2}{c|}{M$\to$M} & \multicolumn{2}{c}{M$\to$D}\\ \cline{2-9}
    & Rank-1    & mAP   & Rank-1 & mAP & Rank-1    & mAP   & Rank-1 & mAP \\ \toprule[1pt]
    $2s$ &  93.3   &  81.9  & 50.0 & 31.6 &  93.7   &  81.6  & 50.3 & 31.9  \\
    $2c$ & \textbf{93.9}   &  82.2  & 50.6 & 32.0 &  93.7   &  81.6  & \textbf{52.2} & 32.7  \\
    $2h$ &  93.5   &  82.3  & 50.3 & 31.8 &   93.5  &  81.6  & 51.0 &  32.1 \\
    $2cs$ &   93.5  &  81.6  & 49.4 & 31.3 &  93.6   &  81.5  & 52.0 & \textbf{33.0}  \\
    $2sc$ &  93.4   &  81.7  & 49.5 & 30.9 &  93.4   &  81.6  & 50.7 &  32.4 \\
    $2sum$ &  93.4   & \textbf{82.4}   & \textbf{52.0} & \textbf{32.5} &   93.5  &  \textbf{82.0}  & 50.2 & 31.7  \\
    $2multiply$ &  93.4   &  82.1  & 49.7 & 31.1 &  \textbf{93.8}   &  81.6  & 49.9 & 31.8  \\
    \toprule[2pt]
  \end{tabular}
\end{table*}

We use the backbone network introduced in Sec. \ref{ssec:network} and those training details in Sec. \ref{ssec:implement} as our baseline. It mainly includes the following components:
\begin{itemize}
\item Input image is resized to $384 \times 128$ compared to usually used $256 \times 128$.
\item Loss function: softmax+triplet combination vs only softmax or triplet.
\item Fixing the backbone network for 5 epochs or not.
\item Employing the warmup strategy or not.
\item Employing the batch normalization ($BN$) layer after the final pooling layer or not.
\end{itemize}

We mainly conducted experiments training only on Market-1501 dataset and testing on both Market-1501 and DukeMTMC-reID datasets, trying to directly exploit our model trained only on the source-domain data (Market-1501) to perform the cross-domain (from Market-1501 to DukeMTMC-reID) Re-ID task. The dimension of final person features is set $1024$. The corresponding results are listed in Table \ref{tab:baseline}, from which, we can know that.
\begin{enumerate}[(a)]
\item All of these components, including both of the architecture designs and the training details, utilized in the backbone network truly can improve the performance Re-ID.
\item Compared to some recently published methods, we give a extremely strong baseline. Our baseline is even better than those methods with multi-layer features fusion (DaRe \cite{wang2018resource}), multi region partition (PCB+RPP \cite{sun2018beyond}), image generating (CamStyle \cite{zhong2018camera}), and attention mechanism (MLFN \cite{chang2018multi}, HACNN \cite{chang2018multi} and Mancs \cite{wang2018mancs}).
\end{enumerate}

\subsection{The effectiveness of attention modules incorporation}
\label{ssec:ex_attn}

As described in Sec. \ref{ssec:atten_incor}, we consider the attention modules incorporation following 3 cases, $attn2$, $attn3$ and $attn4$. For the two type of attention mechanisms, type I: long-range dependency based attention and type II: direct generation based attention described in Sec. \ref{sec:attn}, we respectively conducted experiments training on Market-1501 dataset, testing on Market-1501 and DukeMTMC-reID datasets, to verify the effectiveness of these attention modules. Since the type I attention modules consist of many learnable parameters which leads them hard to converge in the case of $attn4$ under our training settings, we evaluated them only in case $attn2$ and $attn3$, while test type II attention modules in case $attn3$ and $attn4$.
For the batch attention module in type I, which mainly depends on sample compositions of every testing mini-batch (as analyzed in the supplementary materials), we conducted experiments for 10 times changing the mini-batch size in range [10:10:100], recording the mean and std variance.
The results are reported in Table \ref{tab:attn}, from which we can find that.
\begin{enumerate}[(a)]
\item Compared to our baseline in Table \ref{tab:baseline}, surprising results are obtained. (1) Those attention modules incorporation methods perform not better in the single-domain training-testing setting (M$\to$M), which demonstrates the effectiveness of our baseline for single-domain person Re-ID. (2) \textbf{However, those attention modules always achieve much better performance in the cross-domain setting (M$\to$D) with large margins. It demonstrates the effectiveness of attention incorporation for model generalization and adaptation.}
\item For type I: long-range dependency based attention, (1) in the single-domain setting, $attn2$ and $attn3$ obtain comparable performance. However in the cross-domain setting, $attn3$ always performs much better than $attn2$, which maybe attribute to that the learned features from $attn3$ containing higher semantical information of person body compared to those from $attn2$.
    (2) In the $attn3$ case under cross-domain setting, the spatial and channel combination attention modules seem to perform better than those spatial or channel attention alone.
    (3) The batch attention is truly with randomness depending on the composition of testing mini-batch samples.
\item For type II: direct generation based attention, (1) in the single-domain setting, $attn3$ and $attn4$ also obtain comparable performance. However, in the cross-domain setting, $attn3$ and $attn4$ performs differently, maybe better, maybe worse than the other one. There is no regular to follow.
    (2) For those attention modules, spatial or channel attention alone, or their different arrangements, all of them also perform differently under different settings, not always better or worse than others.
\item In all cases, the channel attention module performs the same or a little better compared to the spatial attention module.
\item In summary, Type I: long-range dependency based attention under $attn3$ case, performs the best in cross-domain setting (M$\to$D, 54.8 ($1sum$) vs. 48.9 (baseline) at Rank-1 (\%)).  It demonstrates the effectiveness of computing attention in the long-range dependency manner for extracting discriminative person features.
\end{enumerate}

\subsection{The effectiveness of additional attention features incorporation}
\label{ssec:ex_attn_skip}
In the above section, we evaluate the effectiveness of attention modules incorporation, especially in the cross-domain setting. In this section, we consider to additionally incorporate the attention features, forming our final attention-based discriminative feature learning (ADFL) method. Based on the experimental results in the above section \ref{ssec:ex_attn}, we will adopt the attention modules in the following cases.
For attention modules, the type I attention in case $attn3$ with spatial and channel attention combination methods $sum$, $cs$ and $sc$ will be adopted.
For attention features incorporation, we mainly consider $AF3$ under concatenation ($cat$) and summation ($sum$) fusion methods, with dimensions of $512$, $1024$ and $2048$. As illustrated in Fig. \ref{fig:framework}, attention features are passed into a $conv block$, then fused with other features. 
The details of the combination methods and the corresponding results are listed in Table \ref{tab:amfi}, where the methods $A-B-C-D-E$ denotes ``the attention features $-$ the fusion method $-$ the dimension outputted from the $conv block$ $-$ the position of attention modules incorporation $-$ the attention modules''.
From the results we can find that.
\begin{table}
\caption{Evaluating the effectiveness of our proposed attention-based mid-level features incorporation under different settings by listing the results in the form of Rank-1(mAP) (\%).}
\label{tab:amfi}
  \centering
  \begin{tabular}{l|c|c}
    \toprule[2pt]
    Method  & M$\to$M & M$\to$D \\
    \toprule[1pt]
     baseline (ours) & 94.0 (83.1) & 48.9 (30.8)   \\
     \hline
     $AF3-cat-512-attn3-1sum$ & 94.0 (83.4) & 53.5 (34.3)   \\
     $AF3-cat-512-attn3-1sc$ & 94.2 (\textbf{84.4}) & 53.6 (34.7)   \\
     $AF3-cat-512-attn3-1cs$ & 94.1 (83.6) & 54.0 (34.9)   \\
     \hline
     $AF3-cat-1024-attn3-1sum$ & \textbf{94.4} (83.1) & \textbf{56.8} (\textbf{37.4})   \\
     $AF3-cat-1024-attn3-1sc$ & 94.1 (83.2) & 53.3 (34.6)   \\
     $AF3-cat-1024-attn3-1cs$ & 93.6 (82.8) & 54.5 (36.0)   \\
     \hline
     $AF3-sum-1024-attn3-1sum$ & 94.1 (81.8) & 56.1 (36.4)   \\
     $AF3-sum-1024-attn3-1sc$ & 93.9 (82.6) & 54.3 (34.9)   \\
     $AF3-sum-1024-attn3-1cs$ & 93.8 (81.8) & 54.7 (36.1)   \\
     \hline
     $AF3-sum-2048-attn3-1sum$ & 94.0 (82.1) & 56.0 (36.6)   \\
     $AF3-sum-2048-attn3-1sc$ & 94.0 (82.4) & 55.2 (35.1)    \\
     $AF3-sum-2048-attn3-1cs$ & 93.7 (81.5) & 55.4 (36.3)   \\
    \toprule[2pt]
  \end{tabular}
\end{table}

\begin{enumerate}[(a)]
\item In single-domain setting, those ADFL methods obtain comparable results compared to the baseline method. However, in the cross-domain setting those ADFL methods perform much better than the baseline method. M$\to$D, 56.8 ($AF3-cat-1024-attn3-1sum$) vs. 48.9 (baseline) at Rank-1 (\%), there are almost 8 percent improvements. 
    Therefore, in the following we will only focus on the cross-domain setting.
\item Compared those results in Table \ref{tab:attn} with Table \ref{tab:amfi}, we can find that the ADFL methods, which combine the attention modules and the attention features, achieve better performance compared to the corresponding methods with only attention modules. It demonstrates the effectiveness of our proposed attention-based discriminative feature learning for enhancing the model generalization and adaptation by improving the person features with discrimination and robustness.
\item For the attention modules, the $1sum$ methods always perform better than $1sc$ and $1cs$ except in the $512$ cases.
\item $512$ methods perform worse compared to $1024$ and $2048$ methods.
\item In $attn3-1sum$ cases, $AF3-cat-1024$, $AF3-sum-1024$ and $AF3-sum-2048$ can achieve comparable performance.
\end{enumerate}
\subsection{Comparison with state-of-the-art}
\label{ssec:ex_sota}

\begin{table*}
\caption{Comparison with state-of-the-art methods on Market1501 and DukeMTMC-reID, under single-domain setting, by listing the Rank-1 and mAP results (\%).}
\label{tab:sota_single}
  \centering
  \begin{tabular}{l|c|c|c|c|c}
    \toprule[2pt]
    \multirow{2}{*}{Method} & \multirow{2}{*}{Publication} & \multicolumn{2}{c|}{Market-1501} & \multicolumn{2}{c}{DukeMTMC-reID} \\
     \cline{3-6}
                    &  & Rank-1    & mAP   & Rank-1 & mAP    \\ \toprule[1pt]
     SVDNet \cite{sun2017svdnet}  & ICCV17 & 82.3 & 62.1 & 76.7 & 56.8   \\
     PDC \cite{su2017pose}  & ICCV17 & 84.4 & 63.4 & - & -   \\
     PAR \cite{zhao2017deeply}  & ICCV17 & 81.0 & 63.4 & - & -   \\
     GLAD \cite{wei2017glad} & MM17 & 89.9 & 73.9 & - & -   \\
     JLML \cite{li2017person} & IJCAI17 & 85.1 & 65.5 & - & -   \\
     TripletLoss \cite{hermans2017defense} & arXiv17 & 84.9 & 69.1 & - & -   \\
     CDIM \cite{yu2017devil} & arXiv17 & 89.9 & 75.6 & 80.4 & 63.9   \\
     AlignedReID \cite{zhang2017alignedreid} & arXiv17 & 92.6 & 82.3 & - & -   \\
     DML \cite{zhang2018deep} & CVPR18 & 87.7 & 68.8 & - & -   \\
     DaRe \cite{wang2018resource} & CVPR18 & 86.4 & 69.3 & 75.2 & 57.4   \\
     AOS \cite{huang2018adversarially} & CVPR18 & 86.5 & 70.4 & 79.2 & 62.1   \\
     CamStyle \cite{zhong2018camera} & CVPR18 & 88.1 & 68.7 & 75.3 & 53.5   \\
     MLFN \cite{chang2018multi} & CVPR18 & 90.0 & 68.7 & 81.0 & 62.8   \\
     PSE \cite{sarfraz2018pose} & CVPR18 & 87.7 & 69.0 & 79.8 & 62.0   \\
     SPReID \cite{kalayeh2018human} & CVPR18 & 92.5 & 81.3 & 84.4 & 71.0   \\
     HA-CNN \cite{li2018harmonious} & CVPR18 & 91.2 & 75.7 & 80.5 & 63.8   \\
     CA$^3$Net \cite{liu20183} & MM18 & 93.2 & 80.0 & 84.6 & 70.2 \\
     PNGAN \cite{qian2018pose} & ECCV18 & 89.4 & 72.6 & 73.6 & 53.2   \\
     PABR \cite{suh2018part} & ECCV18 & 91.7 & 79.6 & 84.4 & 69.3   \\
     MANCS \cite{wang2018mancs} & ECCV18 & 93.1 & 82.3 & 84.9 & 71.8   \\
     PCB+RPP \cite{sun2018beyond} & ECCV18 & 93.8 & 81.6 & 83.3 & 69.2   \\

     \hline
     baseline  & ours & 94.0 & \textbf{83.1} & 85.2 & \textbf{72.8}   \\
     \hline
     $AF3-cat-1024-attn3-1sum$& ours  & \textbf{94.4} & \textbf{83.1} & \textbf{86.0} &  \textbf{72.8}  \\
     $AF3-sum-1024-attn3-1sum$& ours  & 94.1 & 81.8 & 85.4 &  71.3  \\
     $AF3-sum-2048-attn3-1sum$& ours  & 94.0 & 82.1 & 85.3 & 70.7   \\
    \toprule[2pt]
  \end{tabular}
\end{table*}

\begin{table*}
\caption{Comparison with state-of-the-art methods on Market1501 and DukeMTMC-reID, under cross-domain setting, by listing the Rank-1 and mAP results (\%). Label M and D denote Market-1501 and DukeMTMC-reID, respectively. M$\to$D means training on M and testing on D.}
\label{tab:sota_cross}
  \centering
  \begin{tabular}{l|c|c|c|c|c}
    \toprule[2pt]
    \multirow{2}{*}{Method} & \multirow{2}{*}{Publication} & \multicolumn{2}{c|}{M$\to$D} & \multicolumn{2}{c}{D$\to$M} \\
     \cline{3-6}
                    &  & Rank-1    & mAP   & Rank-1 & mAP    \\ \toprule[1pt]
     PUL \cite{fan2018unsupervised}  & TOMM18 & 30.0 & 16.4 & 45.5 & 20.5   \\
     PTGAN \cite{wei2018person}  & CVPR18 & 27.4 & - & 38.6 & -   \\
     SPGAN \cite{deng2018image}  & CVPR18 & 41.1 & 22.3 & 51.5 & 22.8   \\
     SPGAN+LMP \cite{deng2018image}  & CVPR18 & 46.4 & 26.2 & 57.7 & 26.7   \\
     TJ-AIDL \cite{wang2018transferable}  & CVPR18 & 44.3 & 23.0 & 58.2 & 26.5   \\
     HHL \cite{zhong2018generalizing}  & ECCV18 & 46.9 & 27.2 & 62.2 & 31.4   \\
     PCB \cite{sun2018beyond}  & ECCV18 & 42.9 & 23.8 & 56.5 & 27.7   \\
     PAP-S-PS \cite{huang2018eanet} & arXiv18 & 51.4 & 31.7 & 61.7 & 32.9   \\
     PAP-ST-PS \cite{huang2018eanet} & arXiv18 & 56.4 & 36.0 & 66.1 & 35.8   \\

     \hline
     baseline  & ours & 48.9 & 30.8  & 59.4 &  29.1  \\
     \hline
     $AF3-cat-1024-attn3-1sum$& ours  & \textbf{56.8} & \textbf{37.4} & \textbf{67.2} &  \textbf{36.3}  \\
     $AF3-sum-1024-attn3-1sum$& ours  & 56.1 & 36.4 & 65.6 &  35.2  \\
     $AF3-sum-2048-attn3-1sum$& ours  & 56.0 & 36.6 & 66.6 & 35.0   \\
    \toprule[2pt]
  \end{tabular}
\end{table*}

\begin{table*}
\caption{The results of proposed models trained on MSMT17 dataset, by listing the Rank-1 and mAP results (\%). Label M, D and MS denote Market-1501, DukeMTMC-reID and MSMT17 respectively. MS$\to$D means training on MS and testing on D.}
\label{tab:sota_msmt17}
  \centering
  \begin{tabular}{l|c|c|c|c|c|c}
    \toprule[2pt]
    \multirow{2}{*}{Method}  & \multicolumn{2}{c|}{MS$\to$MS} & \multicolumn{2}{c|}{MS$\to$M} & \multicolumn{2}{c}{MS$\to$D} \\
    \cline{2-7}
                    & Rank-1    & mAP   & Rank-1 & mAP & Rank-1 & mAP   \\ \toprule[1pt]
     GoogleNet \cite{szegedy2015going}  &  47.6 & 23.0 & - & - & - & -   \\
     PDC \cite{su2017pose}  &  58.0 & 29.7 & - & - & - & -   \\
     GLAD \cite{wei2017glad}  &  61.4 & 34.0 & - & - & - & -   \\
     \hline
     baseline  & 76.0  &  48.5 & 62.1 & 33.0 & 63.7 & 43.3  \\
     \hline
     $AF3-cat-1024-attn3-1sum$& \textbf{78.2}  & \textbf{48.8} & \textbf{68.0} & \textbf{37.7}  & \textbf{66.3} & \textbf{46.2}  \\
     $AF3-sum-1024-attn3-1sum$& 74.5  & 45.7 & 61.6  & 32.4 & 62.1 & 41.5   \\
     $AF3-sum-2048-attn3-1sum$&  76.8 & 46.7 & 67.4 & 36.7 & 66.1 &  45.8  \\
    \toprule[2pt]
  \end{tabular}
\end{table*}

We compare our proposed ADFL methods with state-of-the-art methods on Market-1501, DukeMTMC-reID and MSMT17 datasets, under both single-domain and cross-domain settings. From the results in Table \ref{tab:amfi}, we decide to adopt $AF3-cat-1024-attn3-1sum$, $AF3-sum-1024-attn3-1sum$ and $AF3-sum-2048-attn3-1sum$ as the delegates of our ADFL methods.
Table \ref{tab:sota_single} lists the results of Market-1501 and DukeMTMC-reID under single-domain setting.
Table \ref{tab:sota_cross} lists the results of Market-1501 and DukeMTMC-reID under cross-domain setting.
Table \ref{tab:sota_msmt17} lists the corresponding results of MSMT17 under both single-domain and cross-domain settings.

From the results in those Tables, we can find that.
\begin{enumerate}[(a)]
\item We implement a extreme strong baseline, which may be much better than some state-of-the-art methods, attributing to the empirical architecture modifications and training strategies. This situation is especially obvious in single-domain setting, as shown in Table \ref{tab:sota_single} and \ref{tab:sota_msmt17}.
\item The attention-based discriminative feature learning method is effective for enhancing the model generalization and adaptation, which especially could be verified in the cross-domain setting by the results in Table \ref{tab:sota_cross}. Our ADFL methods achieve the best performance, even better than those methods which leverage the information of target-domain data. However, we perform the cross-domain person Re-ID by directly exploiting our ADFL model training only on the source-domain without any auxiliary information.
\item Compared to those existing cross-domain person Re-ID methods, which leverage the information of target-domain data and perform more tasks \cite{wang2018transferable,wei2018person,deng2018image,huang2018eanet,zhong2018generalizing}, the big success of our simple attention incorporation for cross-domain person Re-ID can direct us a way to effectively design models for cross-domain tasks.
\end{enumerate}

\section{Conclusions}
\label{sec:conc}
 This work mainly aims at enhancing the model generation and adaptation, by focusing on adaptively extracting the discriminative and robust person features. We emphasized to learn discriminative person features, through the simple incorporation of attention modules. It can improve the learned person features with more discrimination and robustness. Based on our strong baseline and attention modules incorporation, the experimental results on three Re-ID datasets demonstrated the effectiveness of our proposed ADFL methods compared to the state-of-the-art approaches, especially under the cross-domain setting by directly exploiting a pre-trained Re-ID model to new domains. The surprisingly good results by only simple attention incorporation may give us some new insights when consider the cross-domain tasks in the future.

 \section*{Acknowledgments}
This work was supported by the National Natural Science Foundation of China (grant numbers 61671125, 61201271, 61301269), and the State Key Laboratory of Synthetical Automation for Process Industries (grant number PAL-N201401).

\bibliographystyle{IEEEtrans}
\bibliography{reid}

\begin{thebibliography}{10}
\providecommand{\url}[1]{#1}
\csname url@samestyle\endcsname
\providecommand{\newblock}{\relax}
\providecommand{\bibinfo}[2]{#2}
\providecommand{\BIBentrySTDinterwordspacing}{\spaceskip=0pt\relax}
\providecommand{\BIBentryALTinterwordstretchfactor}{4}
\providecommand{\BIBentryALTinterwordspacing}{\spaceskip=\fontdimen2\font plus
\BIBentryALTinterwordstretchfactor\fontdimen3\font minus
  \fontdimen4\font\relax}
\providecommand{\BIBforeignlanguage}[2]{{%
\expandafter\ifx\csname l@#1\endcsname\relax
\typeout{** WARNING: IEEEtranS.bst: No hyphenation pattern has been}%
\typeout{** loaded for the language `#1'. Using the pattern for}%
\typeout{** the default language instead.}%
\else
\language=\csname l@#1\endcsname
\fi
#2}}
\providecommand{\BIBdecl}{\relax}
\BIBdecl

\bibitem{bai2017scalable}
S.~Bai, X.~Bai, and Q.~Tian, ``Scalable person re-identification on supervised
  smoothed manifold,'' in \emph{CVPR}, 2017, pp. 3356--3365.

\bibitem{cao2017realtime}
Z.~Cao, T.~Simon, S.-E. Wei, and Y.~Sheikh, ``Realtime multi-person 2d pose
  estimation using part affinity fields,'' in \emph{CVPR}, 2017, pp.
  1302--1310.

\bibitem{chang2018multi}
X.~Chang, T.~M. Hospedales, and T.~Xiang, ``Multi-level factorisation net for
  person re-identification,'' in \emph{CVPR}, 2018.

\bibitem{corbetta2002control}
M.~Corbetta and G.~L. Shulman, ``Control of goal-directed and stimulus-driven
  attention in the brain,'' \emph{Nature reviews neuroscience}, vol.~3, no.~3,
  p. 201, 2002.

\bibitem{deng2018image}
W.~Deng, L.~Zheng, Q.~Ye, G.~Kang, Y.~Yang, and J.~Jiao, ``Image-image domain
  adaptation with preserved self-similarity and domain-dissimilarity for person
  re-identification,'' in \emph{CVPR}, 2018, pp. 994--1003.

\bibitem{ding2018feature}
G.~Ding, S.~Zhang, S.~Khan, Z.~Tang, J.~Zhang, and F.~Porikli, ``Feature
  affinity based pseudo labeling for semi-supervised person
  re-identification,'' \emph{IEEE Transactions on Multimedia}, 2018.

\bibitem{fan2018unsupervised}
H.~Fan, L.~Zheng, C.~Yan, and Y.~Yang, ``Unsupervised person re-identification:
  Clustering and fine-tuning,'' \emph{ACM Transactions on Multimedia Computing,
  Communications, and Applications}, vol.~14, no.~4, p.~83, 2018.

\bibitem{felzenszwalb2008discriminatively}
P.~Felzenszwalb, D.~McAllester, and D.~Ramanan, ``A discriminatively trained,
  multiscale, deformable part model,'' in \emph{CVPR}, 2008, pp. 1--8.

\bibitem{fu2018horizontal}
Y.~Fu, Y.~Wei, Y.~Zhou, H.~Shi, G.~Huang, X.~Wang, Z.~Yao, and T.~Huang,
  ``Horizontal pyramid matching for person re-identification,'' \emph{arXiv
  preprint arXiv:1804.05275}, 2018.

\bibitem{he2016deep}
K.~He, X.~Zhang, S.~Ren, and J.~Sun, ``Deep residual learning for image
  recognition,'' in \emph{CVPR}, 2016, pp. 770--778.

\bibitem{hermans2017defense}
A.~Hermans, L.~Beyer, and B.~Leibe, ``In defense of the triplet loss for person
  re-identification,'' \emph{arXiv preprint arXiv:1703.07737}, 2017.

\bibitem{hochreiter1997long}
S.~Hochreiter and J.~Schmidhuber, ``Long short-term memory,'' \emph{Neural
  computation}, vol.~9, no.~8, pp. 1735--1780, 1997.

\bibitem{hu2018squeeze}
J.~Hu, L.~Shen, and G.~Sun, ``Squeeze-and-excitation networks,'' in
  \emph{CVPR}, 2018, pp. 7132--7141.

\bibitem{huang2018adversarially}
H.~Huang, D.~Li, Z.~Zhang, X.~Chen, and K.~Huang, ``Adversarially occluded
  samples for person re-identification,'' in \emph{CVPR}, 2018, pp. 5098--5107.

\bibitem{huang2018eanet}
H.~Huang, W.~Yang, X.~Chen, X.~Zhao, K.~Huang, J.~Lin, G.~Huang, and D.~Du,
  ``Eanet: Enhancing alignment for cross-domain person re-identification,''
  \emph{arXiv preprint arXiv:1812.11369}, 2018.

\bibitem{itti1998model}
L.~Itti, C.~Koch, and E.~Niebur, ``A model of saliency-based visual attention
  for rapid scene analysis,'' \emph{IEEE Transactions on pattern analysis and
  machine intelligence}, vol.~20, no.~11, pp. 1254--1259, 1998.

\bibitem{jaderberg2015spatial}
M.~Jaderberg, K.~Simonyan, A.~Zisserman \emph{et~al.}, ``Spatial transformer
  networks,'' in \emph{NeurIPS}, 2015, pp. 2017--2025.

\bibitem{kalayeh2018human}
M.~M. Kalayeh, E.~Basaran, M.~Gokmen, M.~E. Kamasak, and M.~Shah, ``Human
  semantic parsing for person re-identification,'' \emph{CVPR}, 2018.

\bibitem{kingma2014adam}
D.~P. Kingma and J.~Ba, ``Adam: A method for stochastic optimization,''
  \emph{arXiv preprint arXiv:1412.6980}, 2014.

\bibitem{larochelle2010learning}
H.~Larochelle and G.~E. Hinton, ``Learning to combine foveal glimpses with a
  third-order boltzmann machine,'' in \emph{NeurIPS}, 2010, pp. 1243--1251.

\bibitem{li2017learning}
D.~Li, X.~Chen, Z.~Zhang, and K.~Huang, ``Learning deep context-aware features
  over body and latent parts for person re-identification,'' in \emph{CVPR},
  2017, pp. 384--393.

\bibitem{li2017person}
W.~Li, X.~Zhu, and S.~Gong, ``Person re-identification by deep joint learning
  of multi-loss classification,'' in \emph{IJCAI}, 2017, pp. 2194--2200.

\bibitem{li2018harmonious}
------, ``Harmonious attention network for person re-identification,'' in
  \emph{CVPR}, 2018, pp. 2285--2294.

\bibitem{liu2017end}
H.~Liu, J.~Feng, M.~Qi, J.~Jiang, and S.~Yan, ``End-to-end comparative
  attention networks for person re-identification,'' \emph{IEEE Transactions on
  Image Processing}, vol.~26, no.~7, pp. 3492--3506, 2017.

\bibitem{liu20183}
J.~Liu, Z.-J. Zha, H.~Xie, Z.~Xiong, and Y.~Zhang, ``Ca$^3$net:
  Contextual-attentional attribute-appearance network for person
  re-identification,'' in \emph{ACM MM}, 2018, pp. 737--745.

\bibitem{liu2017hydraplus}
X.~Liu, H.~Zhao, M.~Tian, L.~Sheng, J.~Shao, S.~Yi, J.~Yan, and X.~Wang,
  ``Hydraplus-net: Attentive deep features for pedestrian analysis,'' in
  \emph{ICCV}, 2017, pp. 350--359.

\bibitem{lv2018unsupervised}
J.~Lv, W.~Chen, Q.~Li, and C.~Yang, ``Unsupervised cross-dataset person
  re-identification by transfer learning of spatial-temporal patterns,'' in
  \emph{CVPR}, 2018, pp. 7948--7956.

\bibitem{qian2018pose}
X.~Qian, Y.~Fu, T.~Xiang, W.~Wang, J.~Qiu, Y.~Wu, Y.-G. Jiang, and X.~Xue,
  ``Pose-normalized image generation for person re-identification,'' in
  \emph{ECCV}, 2018, pp. 650--667.

\bibitem{ren2015faster}
S.~Ren, K.~He, R.~Girshick, and J.~Sun, ``Faster r-cnn: Towards real-time
  object detection with region proposal networks,'' in \emph{NeurIPS}, 2015,
  pp. 91--99.

\bibitem{rensink2000dynamic}
R.~A. Rensink, ``The dynamic representation of scenes,'' \emph{Visual
  cognition}, vol.~7, no. 1-3, pp. 17--42, 2000.

\bibitem{ristani2016performance}
E.~Ristani, F.~Solera, R.~Zou, R.~Cucchiara, and C.~Tomasi, ``Performance
  measures and a data set for multi-target, multi-camera tracking,'' in
  \emph{ECCV}, 2016, pp. 17--35.

\bibitem{sarfraz2018pose}
M.~S. Sarfraz, A.~Schumann, A.~Eberle, and R.~Stiefelhagen, ``A pose-sensitive
  embedding for person re-identification with expanded cross neighborhood
  re-ranking,'' in \emph{CVPR}, vol.~7, 2018, p.~8.

\bibitem{su2017pose}
C.~Su, J.~Li, S.~Zhang, J.~Xing, W.~Gao, and Q.~Tian, ``Pose-driven deep
  convolutional model for person re-identification,'' in \emph{ICCV}, 2017, pp.
  3980--3989.

\bibitem{suh2018part}
Y.~Suh, J.~Wang, S.~Tang, T.~Mei, and K.~M. Lee, ``Part-aligned bilinear
  representations for person re-identification,'' \emph{ECCV}, 2018.

\bibitem{sun2017svdnet}
Y.~Sun, L.~Zheng, W.~Deng, and S.~Wang, ``Svdnet for pedestrian retrieval,'' in
  \emph{ICCV}, 2017, pp. 3820--3828.

\bibitem{sun2018beyond}
Y.~Sun, L.~Zheng, Y.~Yang, Q.~Tian, and S.~Wang, ``Beyond part models: Person
  retrieval with refined part pooling (and a strong convolutional baseline),''
  in \emph{ECCV}, 2018, pp. 501--518.

\bibitem{szegedy2017inception}
C.~Szegedy, S.~Ioffe, V.~Vanhoucke, and A.~A. Alemi, ``Inception-v4,
  inception-resnet and the impact of residual connections on learning.'' in
  \emph{AAAI}, vol.~4, 2017, p.~12.

\bibitem{szegedy2015going}
C.~Szegedy, W.~Liu, Y.~Jia, P.~Sermanet, S.~Reed, D.~Anguelov, D.~Erhan,
  V.~Vanhoucke, and A.~Rabinovich, ``Going deeper with convolutions,'' in
  \emph{CVPR}, 2015, pp. 1--9.

\bibitem{vaswani2017attention}
A.~Vaswani, N.~Shazeer, N.~Parmar, J.~Uszkoreit, L.~Jones, A.~N. Gomez,
  {\L}.~Kaiser, and I.~Polosukhin, ``Attention is all you need,'' in
  \emph{NeurIPS}, 2017, pp. 5998--6008.

\bibitem{wang2018mancs}
C.~Wang, Q.~Zhang, C.~Huang, W.~Liu, and X.~Wang, ``Mancs: A multi-task
  attentional network with curriculum sampling for person re-identification,''
  in \emph{ECCV}.\hskip 1em plus 0.5em minus 0.4em\relax Springer, 2018, pp.
  384--400.

\bibitem{wang2018learning}
G.~Wang, Y.~Yuan, X.~Chen, J.~Li, and X.~Zhou, ``Learning discriminative
  features with multiple granularities for person re-identification,''
  \emph{ACM MM}, 2018.

\bibitem{wang2018transferable}
J.~Wang, X.~Zhu, S.~Gong, and W.~Li, ``Transferable joint attribute-identity
  deep learning for unsupervised person re-identification,'' \emph{CVPR}, 2018.

\bibitem{wang2018non}
X.~Wang, R.~Girshick, A.~Gupta, and K.~He, ``Non-local neural networks,'' in
  \emph{CVPR}, 2018, pp. 7794--7803.

\bibitem{wang2018resource}
Y.~Wang, L.~Wang, Y.~You, X.~Zou, V.~Chen, S.~Li, G.~Huang, B.~Hariharan, and
  K.~Q. Weinberger, ``Resource aware person re-identification across multiple
  resolutions,'' in \emph{CVPR}, 2018, pp. 8042--8051.

\bibitem{wang2015zero}
Z.~Wang, R.~Hu, C.~Liang, Y.~Yu, J.~Jiang, M.~Ye, J.~Chen, and Q.~Leng,
  ``Zero-shot person re-identification via cross-view consistency,'' \emph{IEEE
  Transactions on Multimedia}, vol.~18, no.~2, pp. 260--272, 2015.

\bibitem{wei2018person}
L.~Wei, S.~Zhang, W.~Gao, and Q.~Tian, ``Person transfer gan to bridge domain
  gap for person re-identification,'' in \emph{CVPR}, 2018, pp. 79--88.

\bibitem{wei2017glad}
L.~Wei, S.~Zhang, H.~Yao, W.~Gao, and Q.~Tian, ``Glad: global-local-alignment
  descriptor for pedestrian retrieval,'' in \emph{ACM MM}, 2017, pp. 420--428.

\bibitem{wei2018glad}
------, ``Glad: Global-local-alignment descriptor for scalable person
  re-identification,'' \emph{IEEE Transactions on Multimedia}, 2018.

\bibitem{xiao2018simple}
B.~Xiao, H.~Wu, and Y.~Wei, ``Simple baselines for human pose estimation and
  tracking,'' \emph{ECCV}, 2018.

\bibitem{xiao2017joint}
T.~Xiao, S.~Li, B.~Wang, L.~Lin, and X.~Wang, ``Joint detection and
  identification feature learning for person search,'' in \emph{CVPR}, 2017,
  pp. 3376--3385.

\bibitem{xiong2018towards}
F.~Xiong, Y.~Xiao, Z.~Cao, K.~Gong, Z.~Fang, and J.~T. Zhou, ``Towards good
  practices on building effective cnn baseline model for person
  re-identification,'' \emph{arXiv preprint arXiv:1807.11042}, 2018.

\bibitem{xu2018attention}
J.~Xu, R.~Zhao, F.~Zhu, H.~Wang, and W.~Ouyang, ``Attention-aware compositional
  network for person re-identification,'' \emph{CVPR}, 2018.

\bibitem{yu2017devil}
Q.~Yu, X.~Chang, Y.-Z. Song, T.~Xiang, and T.~M. Hospedales, ``The devil is in
  the middle: Exploiting mid-level representations for cross-domain instance
  matching,'' \emph{arXiv preprint arXiv:1711.08106}, 2017.

\bibitem{zeiler2014visualizing}
M.~D. Zeiler and R.~Fergus, ``Visualizing and understanding convolutional
  networks,'' in \emph{ECCV}, 2014, pp. 818--833.

\bibitem{zhang2017alignedreid}
X.~Zhang, H.~Luo, X.~Fan, W.~Xiang, Y.~Sun, Q.~Xiao, W.~Jiang, C.~Zhang, and
  J.~Sun, ``Alignedreid: Surpassing human-level performance in person
  re-identification,'' \emph{arXiv preprint arXiv:1711.08184}, 2017.

\bibitem{zhang2018deep}
Y.~Zhang, T.~Xiang, T.~M. Hospedales, and H.~Lu, ``Deep mutual learning,'' in
  \emph{CVPR}, 2018, pp. 4320--4328.

\bibitem{zhao2017spindle}
H.~Zhao, M.~Tian, S.~Sun, J.~Shao, J.~Yan, S.~Yi, X.~Wang, and X.~Tang,
  ``Spindle net: Person re-identification with human body region guided feature
  decomposition and fusion,'' in \emph{CVPR}, 2017, pp. 907--915.

\bibitem{zhao2017deeply}
L.~Zhao, X.~Li, Y.~Zhuang, and J.~Wang, ``Deeply-learned part-aligned
  representations for person re-identification,'' in \emph{ICCV}, 2017, pp.
  3219--3228.

\bibitem{zheng2015scalable}
L.~Zheng, L.~Shen, L.~Tian, S.~Wang, J.~Wang, and Q.~Tian, ``Scalable person
  re-identification: A benchmark,'' in \emph{ICCV}, 2015, pp. 1116--1124.

\bibitem{zheng2016person}
L.~Zheng, Y.~Yang, and A.~G. Hauptmann, ``Person re-identification: Past,
  present and future,'' \emph{arXiv preprint arXiv:1610.02984}, 2016.

\bibitem{zheng2018re}
M.~Zheng, S.~Karanam, Z.~Wu, and R.~J. Radke, ``Re-identification with
  consistent attentive siamese networks,'' \emph{arXiv preprint
  arXiv:1811.07487}, 2018.

\bibitem{zhong2018generalizing}
Z.~Zhong, L.~Zheng, S.~Li, and Y.~Yang, ``Generalizing a person retrieval model
  hetero-and homogeneously,'' in \emph{ECCV}, 2018, pp. 172--188.

\bibitem{zhong2018camera}
Z.~Zhong, L.~Zheng, Z.~Zheng, S.~Li, and Y.~Yang, ``Camera style adaptation for
  person re-identification,'' in \emph{CVPR}, 2018, pp. 5157--5166.

\bibitem{zhu2017CycleGAN}
J.-Y. Zhu, T.~Park, P.~Isola, and A.~A. Efros, ``Unpaired image-to-image
  translation using cycle-consistent adversarial networkss,'' in \emph{ICCV},
  2017.

\end{thebibliography}

\clearpage
\noindent\textbf{Supplementary Materials}

\begin{figure*}
  \centering
  \begin{tabular}{c@{\hspace{5mm}}c}
  \includegraphics[width=90mm]{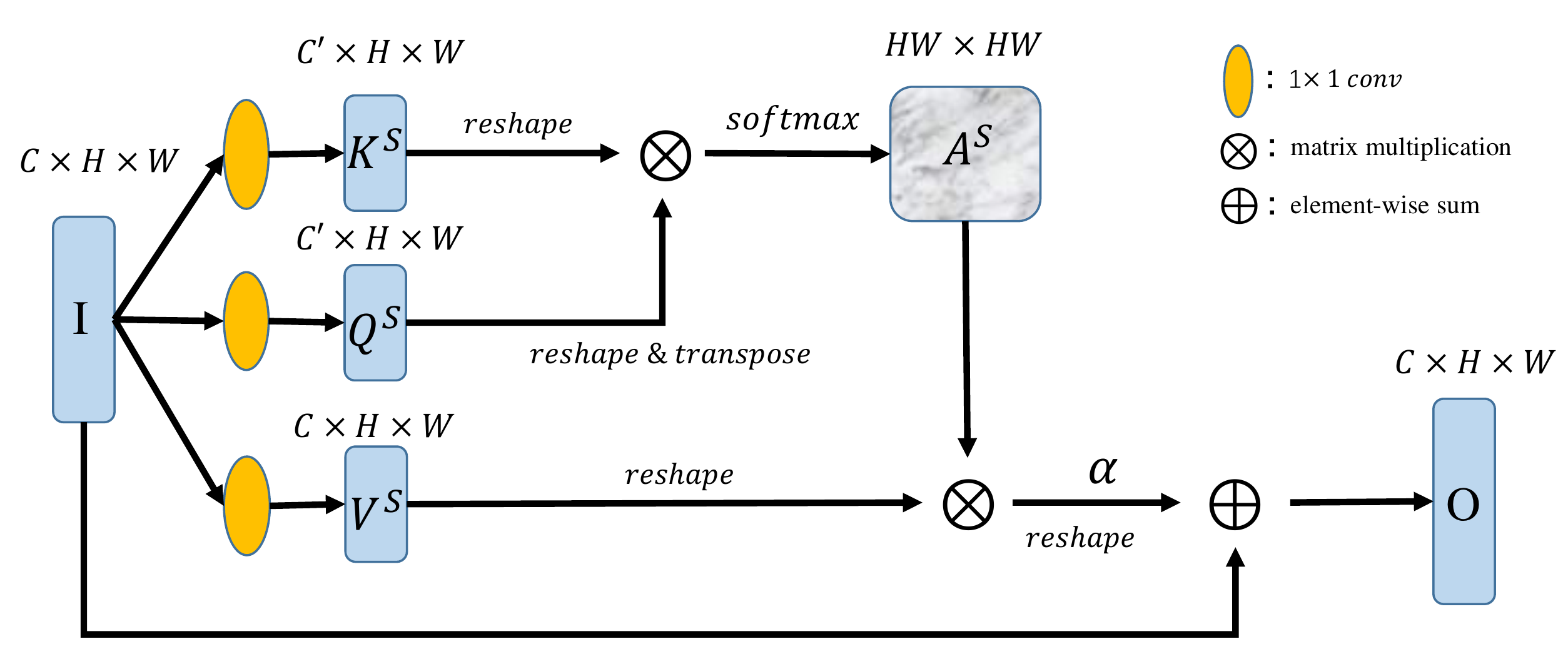} &
  \includegraphics[width=90mm]{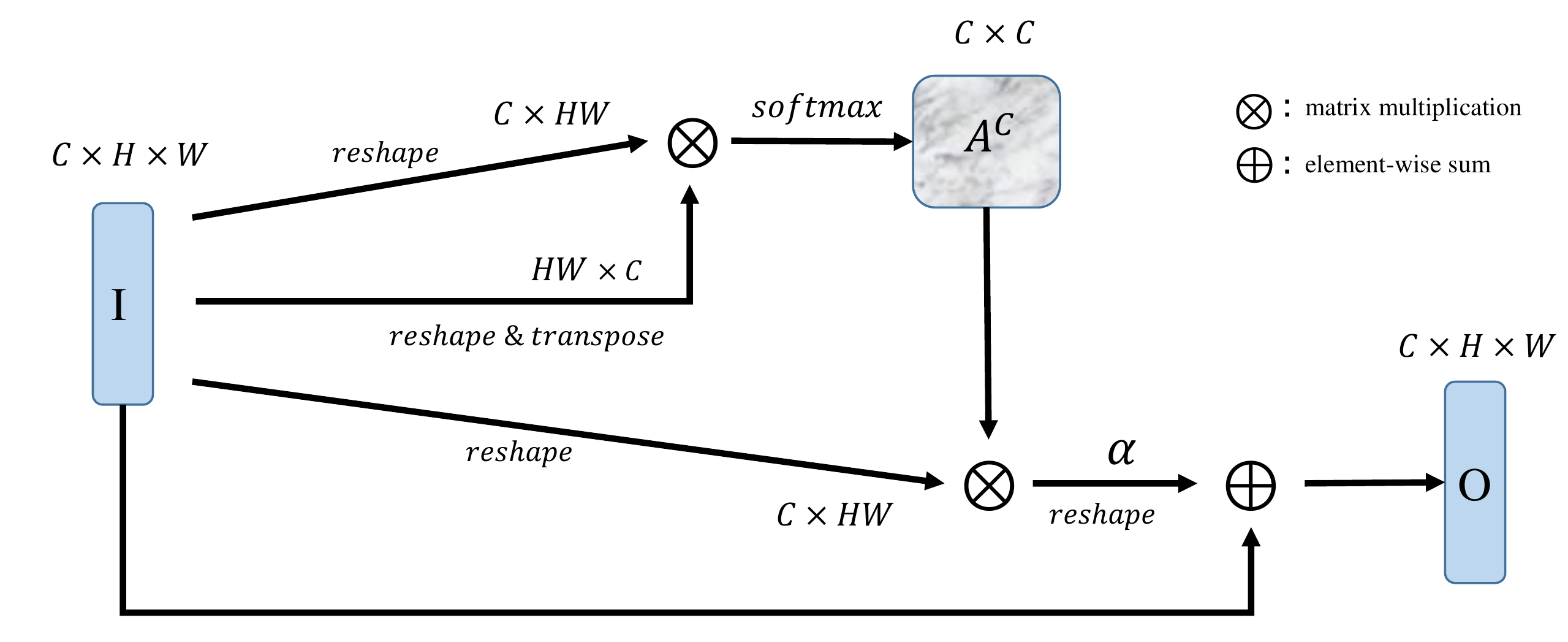}\\
  (a)&(b) \\
  \includegraphics[width=90mm]{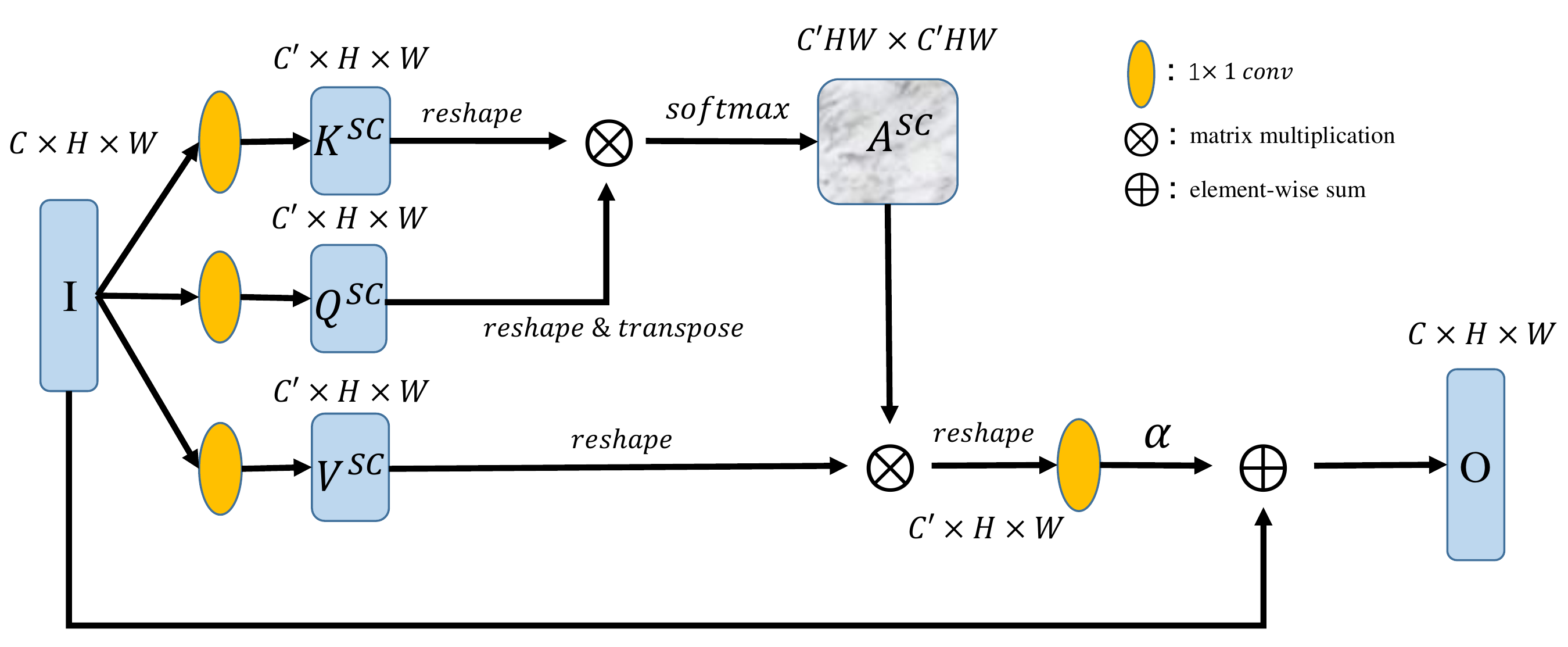} &
  \includegraphics[width=90mm]{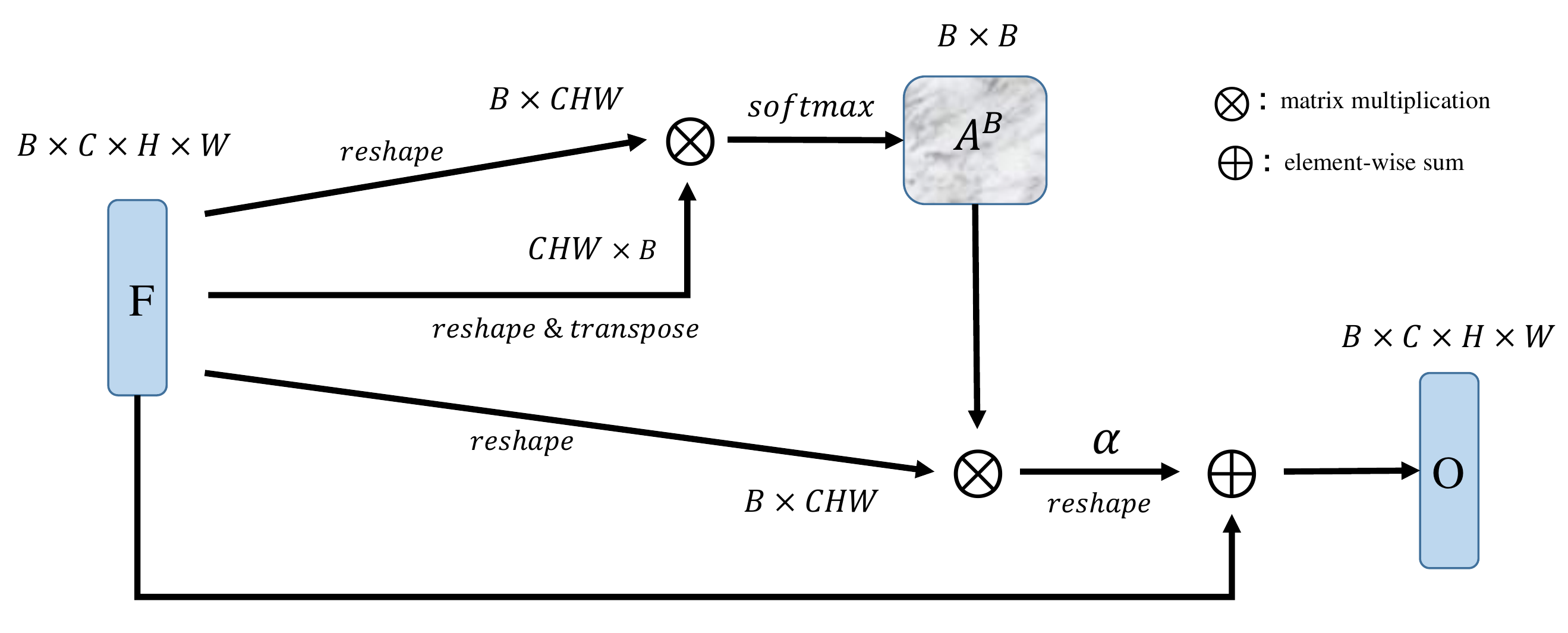}\\
  (c)&(d)
  \end{tabular}
  \caption{Type I: long-range dependency based attention. The details of the attention modules are respectively illustrated in (a) the spatial attention, (b) the channel attention, (c) the hyper attention combining the spatial and channel attention, and (d) the batch attention.}
  \label{fig:attention1}
 \end{figure*}
 
Given an intermediate feature map $F \in R^{B\times C \times H \times W}$ as input, where $B$, $C$, $H$ and $W$ are batch size, channel, height and width, respectively, the structures and computation procedures of different attention modules are as follows.
\subsection{Type I: long-range dependency based attention}
\label{ssec:attn1}
\subsubsection{Spatial attention.}
We accomplish the spatial partition by designing a spatial attention module to model long-range contextual dependencies over local features at all the spatial positions.
The structure of spatial attention is illustrated in Fig. \ref{fig:attention1} (a). The corresponding procedures are as following.
\begin{enumerate}[(i)]
\item Feeding the input feature map of one sample $I \in R^{C \times H \times W}$ into two $1 \times 1 \,\, convolution$ layers to reduce the channel number $C$ to $C'$ (we set $C' = C/8$), generating two new feature maps $K^S \in R^{C' \times H \times W}$ and $Q^S \in R^{C' \times H \times W}$, respectively.
\item $K^S$ and $Q^S$ are $reshaped$ to $R^{C' \times HW}$, where the $HW = H \times W$ is the number of spatial features in this feature map. A $matrix\,\,  multiplication$ is performed between the $transpose$ of $Q^S$ and $K^S$.
\item Applying the $softmax$ operation on each row of the multiplication matrix to calculate the spatial attention map $A^{S} \in R^{HW \times HW}$,
    \begin{align}
        A^{S}_{j,i} = \frac{\exp({K^S_i} \cdot Q^S_j)}{\sum_{i=1}^{HW} \exp({K^S_i} \cdot Q^S_j))},   \label{eq:1sa}
    \end{align}
    where $\cdot$ is the dot product of vectors, $A^{S}_{j,i}$ indicates the extent to which the model attends to the $i^{th}$ spatial position when learning the features of $j^{th}$ spatial position. The more similar feature representations of the two spatial positions, the greater correlation will contribute to each other.
\item Feeding the input feature map $I$ into another $1 \times 1 \,\, convolution$ layer to generate a new feature map $V^S$, and $reshaping$ it to $R^{C \times HW}$. Then a $matrix\,\, multiplication$ is performed between $V^S$ and $A^{S}$, and $reshape$ the multiplication result to $R^{C \times H \times W}$.
\item Finally, we further multiply the output of the spatial attention layer by a scale parameter and add back the input feature map to obtain the final output $O$,
    \begin{align}
        O_j = \alpha \sum_{i=1}^{HW}(A^{S}_{j,i} V^{S}_{i}) + I_j, \label{eq:1sa_out}
    \end{align}
    where $\alpha$ is a learnable parameter, initialized as 0 and gradually learn to a suitable value.
\end{enumerate}

From the above procedures, especially the calculations of Eqns (\ref{eq:1sa}) and (\ref{eq:1sa_out}), we can find that the feature at every spatial position of the final output feature $O$ is a weighted sum of the features at all the spatial positions and the original features. It builds the long-range contextual dependencies to make the spatial attention results with a global contextual view by selectively aggregating contexts from all the spatial positions.

\subsubsection{Channel attention}
\label{sssec:1_ca}
We model the channel attention to enhance the feature representation by explicitly exploit the interdependencies between channel maps. Different from the spatial attention module, we directly compute the channel attention map from the input feature map.
The structure of channel attention is illustrated in Fig. \ref{fig:attention1} (b). The corresponding procedures are as following.
\begin{enumerate}[(i)]
\item The input feature map of one sample $I \in R^{C \times H \times W}$ is directly $reshaped$ to $R^{C \times HW}$. A $matrix\,\,  multiplication$ is performed between $I$ and the $transpose$ of $I$.
\item Applying the $softmax$ operation on each row of the multiplication matrix to calculate the channel attention map $A^{C} \in R^{C \times C}$,
    \begin{align}
        A^{C}_{j,i} = \frac{\exp({I_i} \cdot I_j)}{\sum_{i=1}^{C} \exp({I_i} \cdot I_j)},  \label{eq:1ca}
    \end{align}
    where $A^{C}_{j,i}$ indicates the extent to which the model attends to the $i^{th}$ channel when learning the features of $j^{th}$ channel.
\item Performing the $matrix\,\, multiplication$ between $A^{C}$ and $I$, and $reshape$ the multiplication result to $R^{C \times H \times W}$.
\item Finally, we also further multiply the output of the channel attention layer by a scale parameter and add back the input feature map to obtain the final output $O$,
    \begin{align}
        O_j = \alpha \sum_{i=1}^{C}(A^{C}_{j,i} I_{i}) + I_j, \label{eq:1ca_out}
    \end{align}
    where $\alpha$ is a learnable parameter, initialized as 0 and gradually learn to a suitable value.
\end{enumerate}

From the above procedures, especially the calculations of Eqns (\ref{eq:1ca}) and (\ref{eq:1ca_out}), we can find that the feature at every channel of the final output feature $O$ is a weighted sum of the features of all the channels and the original features. It builds the long-range semantic dependencies to make the channel attention results with body part-dependent aggregation from all the channels.

\subsubsection{Hyper attention}
\label{sssec:1_ha}
In this subsection, we consider to simultaneously model the spatial and channel attention, which we term as hyper attention. The motivation is that each channel map of high level feature maps can be regarded as a class-specific response, and different semantic responses are associated with each other. Therefor, we try to build long-range dependencies for every location to all the other locations in the feature map, including both the spatial positions and channels.
The structure of channel attention is illustrated in Fig. \ref{fig:attention1} (c). The corresponding procedures are as following.
\begin{enumerate}[i)]
\item Feeding the input feature map of one sample $I \in R^{C \times H \times W}$ into two $1 \times 1 \,\, convolution$ layers to reduce the channel number $C$ to $C'$ (we set $C' = C/64$ to reduce the parameters in attention map), generating two new feature maps $K^{SC} \in R^{C' \times H \times W}$ and $Q^{SC} \in R^{C' \times H \times W}$, respectively.
\item $K^{SC}$ and $Q^{SC}$ are $reshaped$ to $R^{C'HW}$, where the $C'HW =C' \times H \times W$ is the number of all the locations in this feature map. A $matrix\,\,  multiplication$ is performed between $K^{SC}$ and the $transpose$ of $Q^{SC}$.
\item Applying the $softmax$ operation on each row of the multiplication matrix to calculate the hyper attention map $A^{SC} \in R^{C'HW \times C'HW}$,
    \begin{align}
        A^{SC}_{j,i} = \frac{\exp({K^{SC}_i} \cdot Q^{SC}_j)}{\sum_{i=1}^{C'HW} \exp({K^{SC}_i} \cdot Q^{SC}_j))},   \label{eq:1sca}
    \end{align}
    where $A^{SC}_{j,i}$ indicates the extent to which the model attends to the $i^{th}$ location when learning the features of $j^{th}$ location (Here the location indicates all the locations across both the spatial positions and channels). The more similar feature representations of the two locations, the greater correlation will contribute to each other.
\item Feeding the input feature map $I$ into another $1 \times 1 \,\, convolution$ layer to generate a new feature map $V^{SC}$, and $reshaping$ it to $R^{C'HW}$. Then a $matrix\,\, multiplication$ is performed between $A^{SC}$ and $V^{SC}$, and $reshape$ the multiplication result to $R^{C' \times H \times W}$.
\item Feeding the result into another $1 \times 1 \,\, convolution$ layer to increase the channel number $C'$ back to $C$, obtaining the hyper attention output with size of $R^{C \times H \times W}$.
\item Finally, we also further multiply the output of the hyper attention layer by a scale parameter $\alpha$ and add back the input feature map to obtain the final output $O$.
\end{enumerate}

From the above procedures, especially the calculations of Eqn (\ref{eq:1sca}), we can find that the feature at every location of the final output feature $O$ is a weighted sum of the features at all the locations and the original features. It takes the spatial and channel attention into consideration simultaneously, but maybe with more parameters in the attention map compared to spatial attention and channel attention ($C'HW \times C'HW$ vs. $HW \times HW$ and $C \times C$, respectively).

\subsubsection{Batch attention}
\label{sssec:1_ba}
Given an intermediate feature map $F \in R^{B\times C \times H \times W}$, it consists of $B$ samples in a mini-batch. Those samples may be with similar features and have impact on each other.  To explore the relationship between those samples in a mini-batch, we try to build the batch attention module.
The structure of channel attention is illustrated in Fig. \ref{fig:attention1} (d). The corresponding procedures are as following.
\begin{enumerate}[(i)]
\item The input feature map of a mini-batch with $B$ samples $F \in R^{B \times C \times H \times W}$ is directly $reshaped$ to $R^{B \times CHW}$. A $matrix\,\,  multiplication$ is performed between $F$ and the $transpose$ of $F$.
\item Applying the $softmax$ operation on each row of the multiplication matrix to calculate the batch attention map $A^{B} \in R^{B \times B}$,
    \begin{align}
        A^{B}_{j,i} = \frac{\exp({F_i} \cdot F_j)}{\sum_{i=1}^{B} \exp({F_i} \cdot F_j)},  \label{eq:1ba}
    \end{align}
    where $A^{B}_{j,i}$ indicates the extent to which the model attends to the $i^{th}$ sample when learning the features of $j^{th}$ sample.
\item Performing the $matrix\,\, multiplication$ between $A^{B}$ and $F$, and $reshape$ the multiplication result to $R^{B \times C \times H \times W}$.
\item Finally, we also further multiply the output of the batch attention layer by a scale parameter and add back the input feature map to obtain the final output $O$,
    \begin{align}
        O_j = \alpha \sum_{i=1}^{B}(A^{B}_{j,i} F_{i}) + F_j, \label{eq:1ba_out}
    \end{align}
    where $\alpha$ is a learnable parameter, initialized as 0 and gradually learn to a suitable value.
\end{enumerate}

From the above procedures, especially the calculations of Eqns (\ref{eq:1ba}) and (\ref{eq:1ba_out}), we can find that the feature of each sample is a weighted sum of the features of all the samples in a mini-batch and the original features. It builds relationships among all the samples. However, (1) the batch attention module may work only if those samples in a mini-batch truly have some effects on each other, heavily depending on the composition of mini-batch samples. (2) There are no learnable parameters in the batch attention module to control the possibly existed relationships. (3) Even if those samples in a mini-batch of training truly have some effects on each other, the relationship of those samples in the testing may be totally different to those in training, too much randomness, which may lead the ineffectiveness of the batch attention.

\subsection{Type II: direct generation based attention}
\label{ssec:attn12}

\begin{figure}
  \centering
  \begin{tabular}{c}
  \includegraphics[width=90mm]{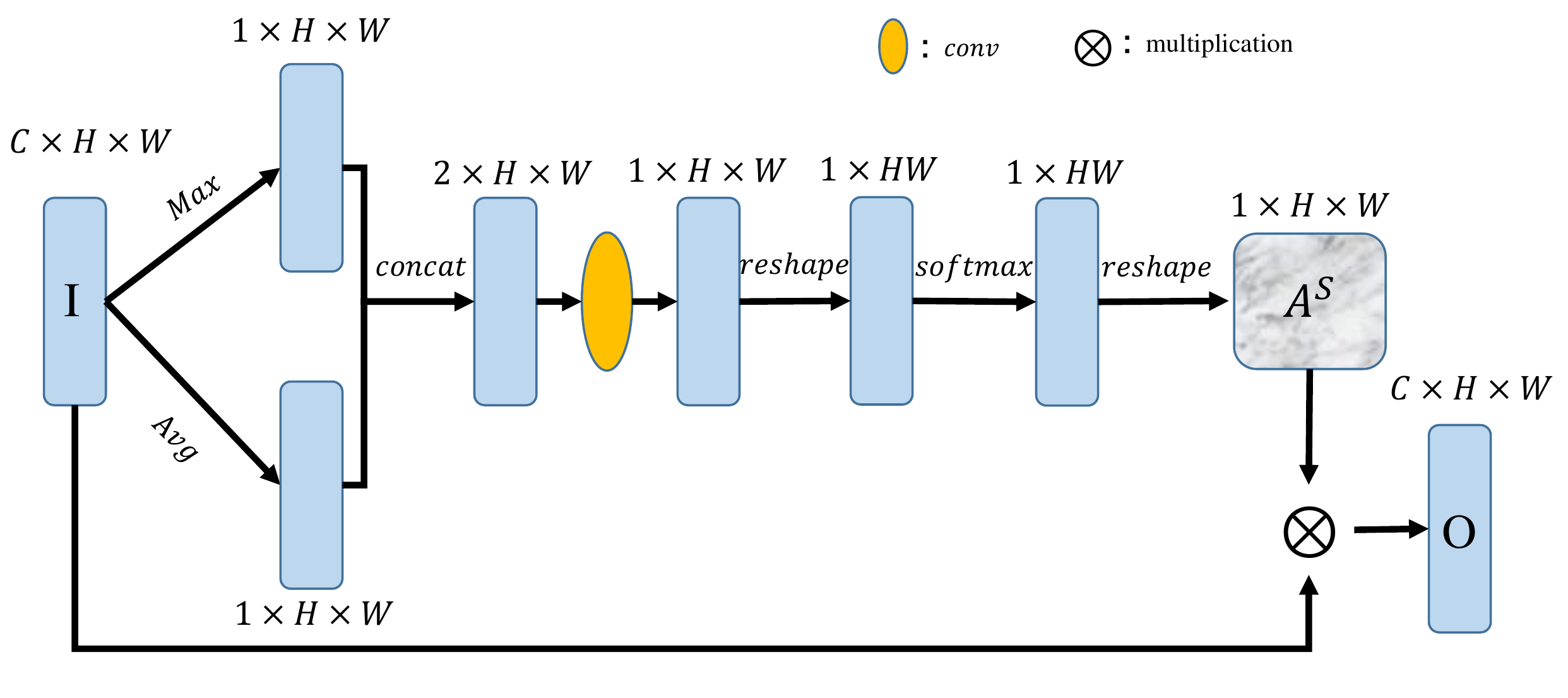}\\   (a) \\
  \includegraphics[width=90mm]{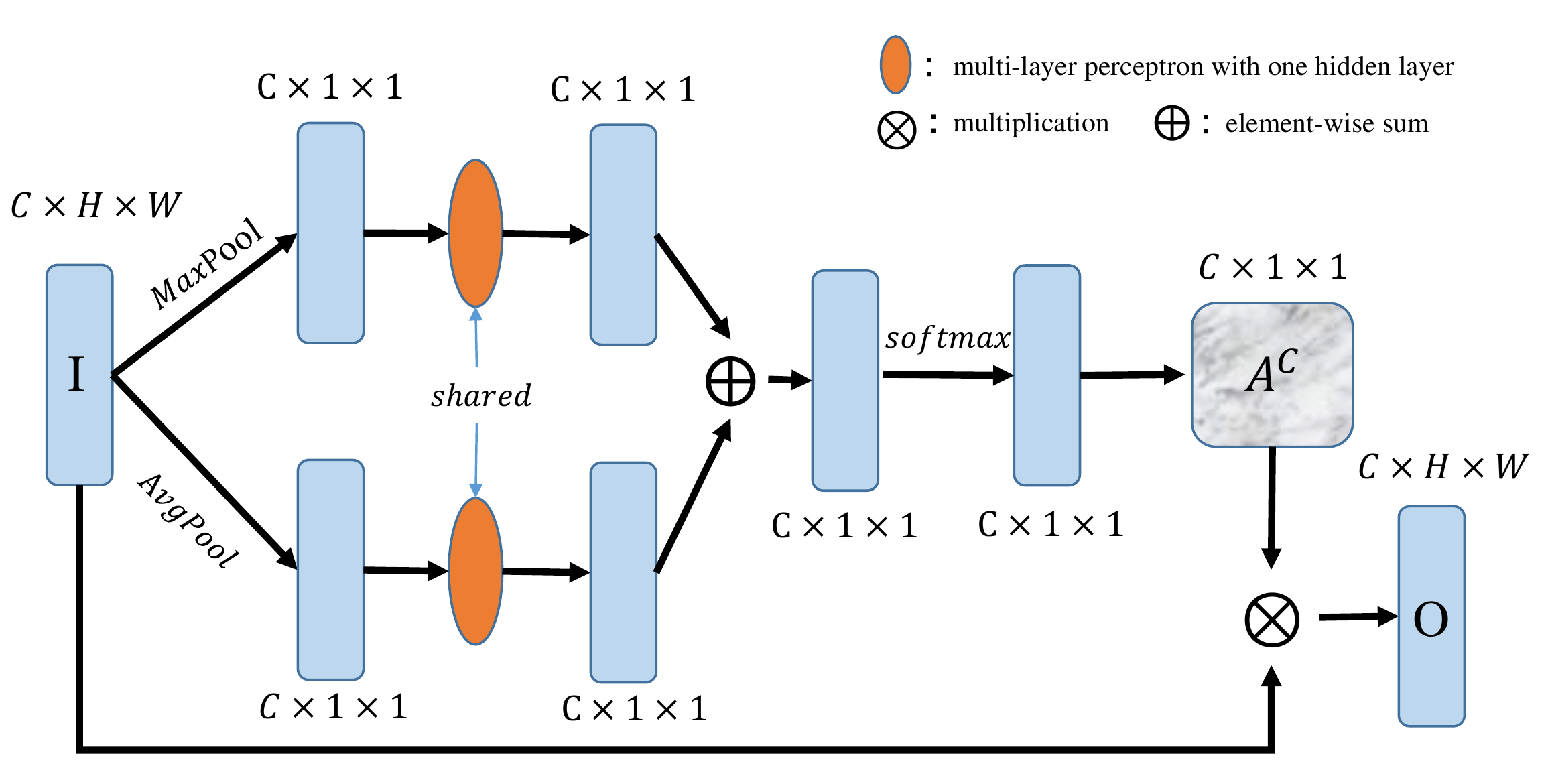}\\   (b)\\
  \includegraphics[width=90mm]{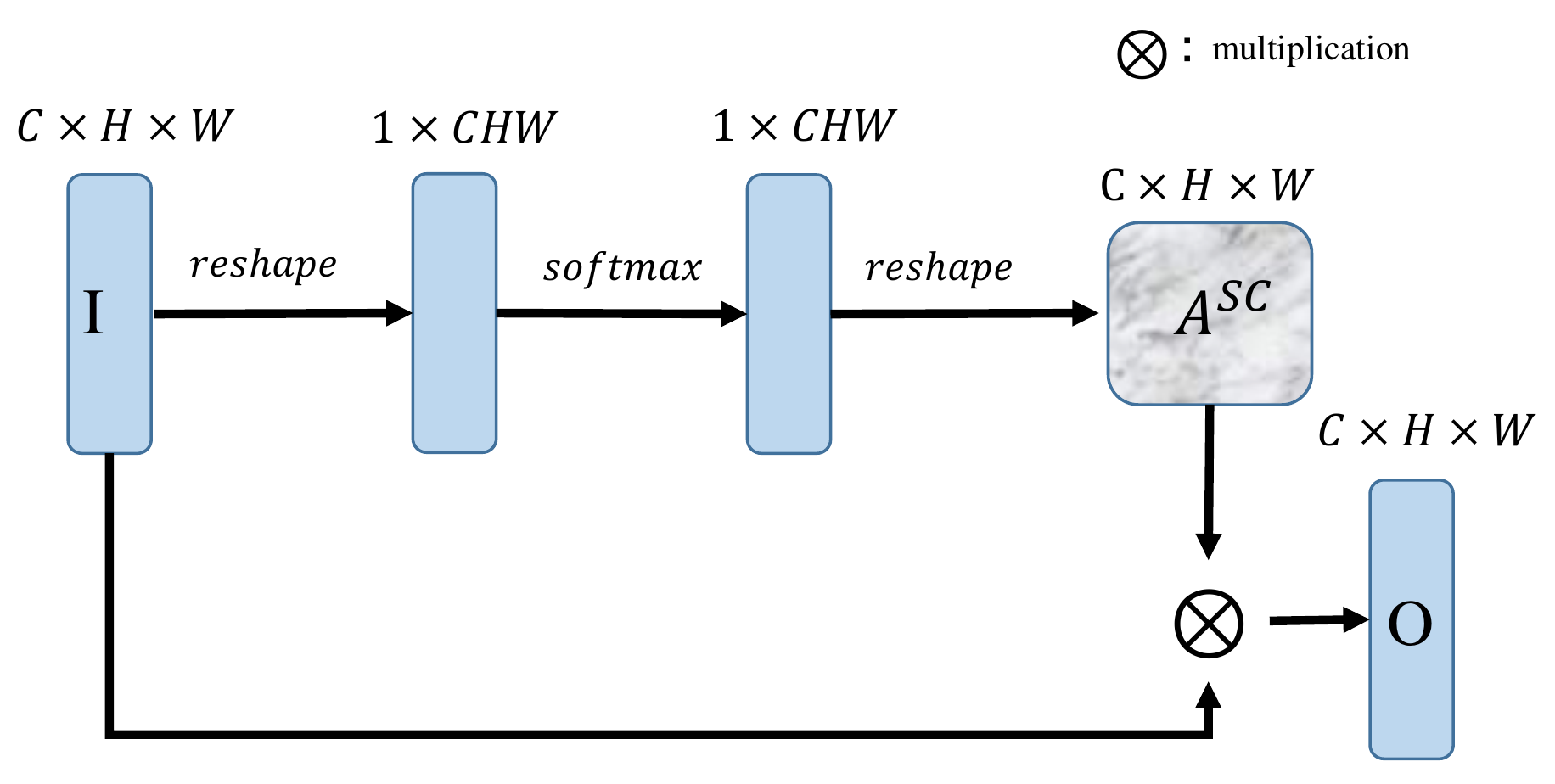}\\  (c)
  \end{tabular}
  \caption{Type II: direct generation based attention. The details of the attention modules are respectively illustrated in (a) the spatial attention, (b) the channel attention, and (c) the hyper attention combining the spatial and channel attention.}
  \label{fig:attention2}
 \end{figure}
 
\subsubsection{Spatial attention.}
We design a spatial attention module to directly generate the spatial attention map by utilizing the inter-spatial relationship of features.
The structure of spatial attention is illustrated in Fig. \ref{fig:attention2} (a). The corresponding procedures are as following.
\begin{enumerate}
\item Given an input feature map of one sample $I \in R^{C \times H \times W}$, performing the $Maximize$ and $Average$ operations along the channel dimension, respectively, to generate two 2D spatial feature maps, then concatenating them to generate a new feature map.
\item Feeding the new feature map to $conv$ layer to reduce the channel number from 2 to 1. The $conv$ layer only change the number of channels, preserving the spatial size $H \times W$. Here we adopt the $3 \times 3 \,\, conv$ layer with $stride = 1$ and $padding = 1$.
\item Then a $softmax$ operation is applied to the 2D spatial feature map, in order to assign each spatial location $(i,j)$ a weight $A^{s}_{i,j}$ indicating the attention importance, we firstly $reshape$ it to a vector and then $reshape$ it back to the 2D spatial feature map.
\item Finally, multiplying the final spatial attention map to the input could obtain the final spatial attention results.
\end{enumerate}

In short, the above procedure can be calculated as follows,
\begin{align}
\begin{tabular}{cl}
    $O$ &$= A^{s} \otimes I$ \\
      &$=  softmax \big(conv([Avg(I);Max(I)])\big)  \otimes I$,
\end{tabular}
\end{align}
where $\otimes$ is the element-wise multiplication. The broadcast mechanism is accordingly performed during multiplication.
\subsubsection{Channel attention}
\label{sssec:2_ca}
In this subsection, we design a channel attention module to directly generate the channel attention map by following the famous channel operation squeeze-and-excitation networks (SE) \cite{hu2018squeeze}. But we additionally utilize the $MaxPool$ combined with $AvgPool$ operation, and the $softmax$ operation to improve the representation power of our channel attention module.
The structure of spatial attention is illustrated in Fig. \ref{fig:attention2} (b). The corresponding procedures are as following.
\begin{enumerate}
\item Performing the squeeze operation. Given an input feature map of one sample $I \in R^{C \times H \times W}$, we adopt $MaxPool$ and $AvgPool$ operations, respectively, to generate two 1D channel feature descriptors.
\item Performing the excitation operation. The two 1D descriptors are then respectively fed into a shared multi-layer perceptron (MLP) network to adaptively re-calibration the two channel feature descriptors. Then sum the the two channel descriptors in element-wise manner.
    The MLP network consists of two $1 \times 1 \,\, conv$ layers with $C/16$ channels as a bottleneck.
\item Applying the $softmax$ operation to the 1D channel feature, in order to assign each channel a weight $A^{C}_{i}$ indicating the degree of attention importance.
\item Finally, multiplying the final channel attention map to the input could obtain the final channel attention results.
\end{enumerate}

In short, the above procedure can be calculated as follows,
\begin{align}
\begin{tabular}{cll}
    $O$ &$= A^{C} \otimes I$ & \\
      &$= softmax\big($ & $MLP(AvgPool(I)) + $ \\
      &  & $MLP(MaxPool(I)) \,\,\big) \,\,  \otimes I$,
\end{tabular}
\end{align}
where $\otimes$ is the element-wise multiplication. The broadcast mechanism is accordingly performed during multiplication.

\subsubsection{Hyper attention}
\label{sssec:2_ha}
In this subsection, we consider to simultaneously model the spatial and channel attentions, which we term as hyper attention.
The structure of spatial attention is illustrated in Fig. \ref{fig:attention2} (c). The corresponding procedures are as following.
\begin{enumerate}
\item Given an input feature map of one sample $I \in R^{C \times H \times W}$, $reshape$ it to $R^{1 \times CHW}$.
\item Applying the $softmax$ operation to the 1D feature descriptors, in order to assign each position a weight $A^{SC}_{i}$ indicating the degree of attention importance.
\item $Reshaping$ it back to $R^{C \times H \times W}$.
\item Finally, multiplying the final hyper attention map to the input could obtain the final channel attention results.
\end{enumerate}

In short, the above procedure can be calculated as follows,
\begin{align}
\begin{tabular}{cl}
    $O$ &$= A^{SC} \otimes I$ \\
        &$= softmax(I) \otimes I $,
\end{tabular}
\end{align}
where $\otimes$ is the element-wise multiplication.

\end{document}